\documentclass[10pt,twocolumn,letterpaper]{article}

\usepackage{cvpr}              
\usepackage{amssymb}
\usepackage{amsmath}
\newcommand{\xmark}{\text{\sffamily X}}

\usepackage{amsmath,amsfonts,bm}

\def\eqref#1{equation~\ref{#1}}

\def\1{\bm{1}}

\def\vc{{\bm{c}}}

\def\vf{{\bm{f}}}

\def\vp{{\bm{p}}}

\def\vs{{\bm{s}}}

\def\mA{{\bm{A}}}

\def\mS{{\bm{S}}}

\def\mW{{\bm{W}}}

\DeclareMathAlphabet{\mathsfit}{\encodingdefault}{\sfdefault}{m}{sl}
\SetMathAlphabet{\mathsfit}{bold}{\encodingdefault}{\sfdefault}{bx}{n}
\newcommand{\tens}[1]{\bm{\mathsfit{#1}}}

\def\sA{{\mathbb{A}}}

\definecolor{cvprblue}{rgb}{0.21,0.49,0.74}
\usepackage[pagebackref,breaklinks,colorlinks,allcolors=cvprblue]{hyperref}
\usepackage{amsmath,amsfonts,bm}
\usepackage{hyperref}
\usepackage{url}
\usepackage{tabularx}
\usepackage{multirow}
\usepackage{booktabs}
\usepackage{listings}
\usepackage{graphicx}
\usepackage{caption}
\usepackage{subcaption}
\usepackage{threeparttable}
\usepackage{listings}
\def\paperID{9850} 
\def\confName{CVPR}
\def\confYear{2025}

\title{Attribute-formed Class-specific Concept Space: Endowing Language Bottleneck Model with Better Interpretability and Scalability}

\author{Jianyang Zhang\textsuperscript{1*},
Qianli Luo\textsuperscript{2*},
Guowu Yang\textsuperscript{1,3},
Wenjing Yang\textsuperscript{4},\\
Weide Liu\textsuperscript{5},
Guosheng Lin\textsuperscript{6}, Fengmao Lv\textsuperscript{2,7,$\dagger$}
\\\\
\textsuperscript{1}University of Electronic Science and Technology of China\qquad\textsuperscript{2}Southwest Jiaotong University \\
\textsuperscript{3}Institute of Electronics and Information Industry Technology of Kash\qquad
\textsuperscript{4}University of Minnesota\\
\textsuperscript{5}Harvard University\qquad\textsuperscript{6}Nanyang Technological University
\\
\textsuperscript{7}Engineering Research Center of Sustainable Urban Intelligent Transportation, Ministry of Education
\\ 
{\tt\small zhang\_jy.1@qq.com},\quad
{\tt\small qianlil@my.swjtu.edu.cn},\quad
{\tt\small guowu@uestc.edu.cn},\quad
{\tt\small wjyang2987@gmail.com},\\
{\tt\small weide001@e.ntu.edu.sg},\quad
{\tt\small gslin@ntu.edu.sg},\quad
{\tt\small fengmaolv@126.com}
}

\usepackage[misc]{ifsym}
\newcommand\blfootnote[1]{
	\begingroup
	\renewcommand\thefootnote{}\footnote{#1}
	\addtocounter{footnote}{-1}
	\endgroup
}
\usepackage[hang]{footmisc}

\begin{document}
\maketitle

\begin{abstract}
	Language Bottleneck Models (LBMs) are proposed to achieve interpretable image recognition by classifying images based on textual concept bottlenecks. However, current LBMs simply list all concepts together as the bottleneck layer, leading to the spurious cue inference problem and cannot generalized to unseen classes. To address these limitations, we propose the Attribute-formed Language Bottleneck Model (ALBM). ALBM organizes concepts in the attribute-formed class-specific space, where concepts are descriptions of specific attributes for specific classes. In this way, ALBM can avoid the spurious cue inference problem by classifying solely based on the essential concepts of each class. In addition, the cross-class unified attribute set also ensures that the concept spaces of different classes have strong correlations, as a result, the learned concept classifier can be easily generalized to unseen classes. Moreover, to further improve interpretability, we propose Visual Attribute Prompt Learning (VAPL) to extract visual features on fine-grained attributes. Furthermore, to avoid labor-intensive concept annotation, we propose the Description, Summary, and Supplement (DSS) strategy to automatically generate high-quality concept sets with a complete and precise attribute. Extensive experiments on 9 widely used few-shot benchmarks demonstrate the interpretability, transferability, and performance of our approach. The code and collected concept sets are available at \url{https://github.com/tiggers23/ALBM}.
	
\end{abstract}

{
	\blfootnote{
		* Equal contribution\\
		$^\dagger$ Corresponding author
	}
}

\section{Introduction}
\label{introduction}
\begin{figure*}[t]
	\centering
	{\includegraphics[width=\linewidth]{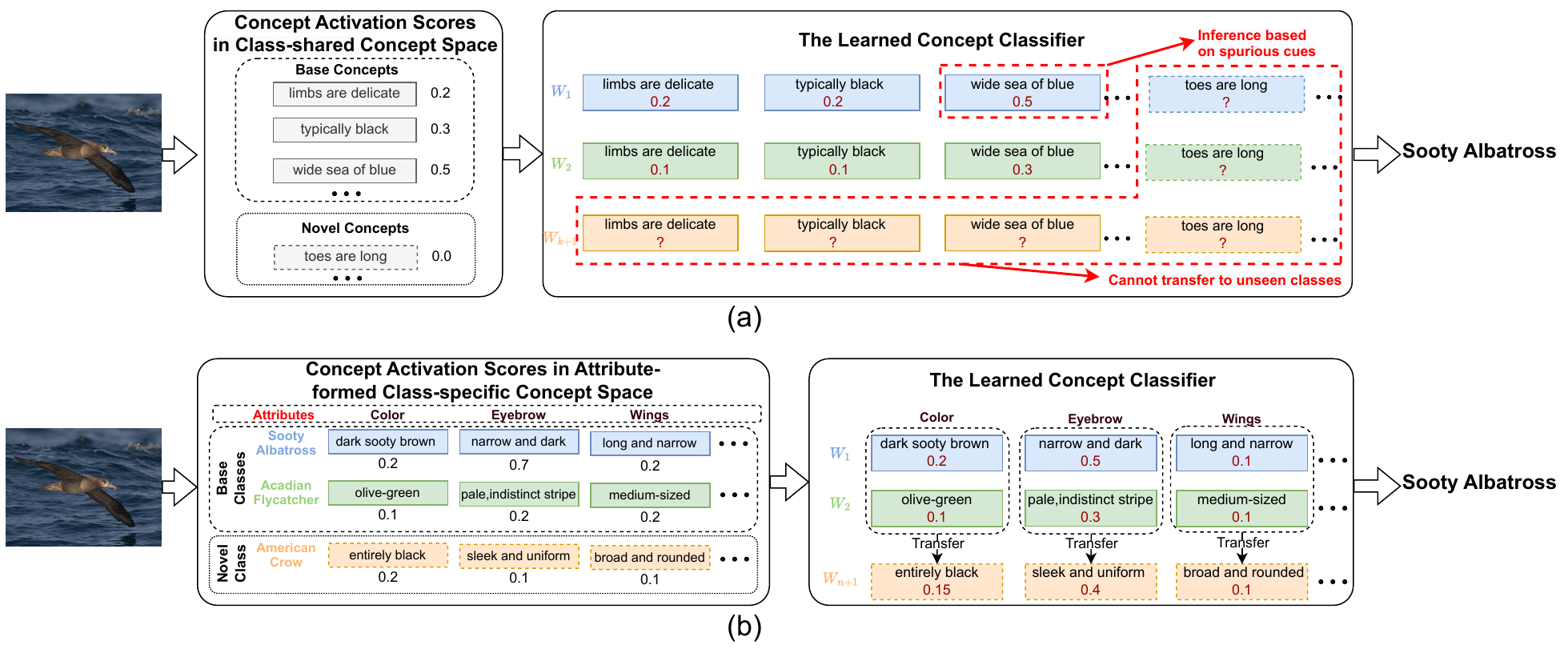}}
	\vspace{-0.7cm}
	\caption{Illustration of the scenario for concept classification. (a) Existing Language Bottleneck Models \cite{DBLP:conf/cvpr/YangPZJCY23, DBLP:conf/iccv/0003WZDHLWSM23} (b) Our Attribute-formed Language Bottleneck Model. Existing LBMs suffer spurious cue inference as they may make decisions based on non-essential or background concepts.
    Additionally, their cross-class scalability is also limited, as expanding the concept space may be necessary for unseen classes.
    On the contrary, our approach predicts classes solely based on their corresponding concepts to avoid the spurious cue problem, and also ensures the cross-category consistent concept space by sharing the unified attribute set, allowing transfer to unseen classes.}
	\label{fig:model}
	\vspace{-0.5cm}
\end{figure*}

Recently, Visual-Language Models (VLMs) \cite{DBLP:conf/icml/RadfordKHRGASAM21, DBLP:conf/icml/JiaYXCPPLSLD21} have shown great performance in visual representation via contrastive pre-training on large-scale internet data. 
To improve the interpretability of VLMs, recent works \cite{DBLP:conf/iccv/0003WZDHLWSM23, DBLP:conf/iccv/PrattCLF23} propose the Training-free Language Bottleneck (TfLB) paradigm, which uses Large Language Model (LLM) \cite{DBLP:conf/nips/BrownMRSKDNSSAA20, DBLP:journals/corr/abs-2307-09288, DBLP:journals/corr/abs-2310-06825} to generate class descriptions in order to construct the concept space, and further achieves interpretable classification by matching images with class concepts.
On this basis, the Language Bottleneck Models (LBMs)  \cite{DBLP:conf/cvpr/YangPZJCY23, DBLP:conf/iclr/YuksekgonulW023, DBLP:conf/iccv/0003WZDHLWSM23, shang2024incremental, kim2023concept} integrate all collected concepts into a unified \textit{class-shared concept space} and train a concept classifier based on this space, achieving both interpretability and discriminability. However, in the class-shared concept space, the concept classifier may learn the relationship between class labels and non-essential concepts co-occurring with the class or concepts from the background (e.g., recognizing a tiger via jungle or recognizing a kind of food via its smell) \cite{DBLP:journals/tmlr/YunBPS23,srivastava2024vlg}, leading to the problem of \textbf{inference based on spurious cues}. Moreover, when generalizing existing models to unseen classes, the concept space may need to be expanded to accommodate the new concepts introduced by these unseen classes. As a result, the concept classifier trained with seen classes \textbf{cannot transfer to unseen classles}. The illustration of the aforementioned two limitations is shown in Fig.~\ref{fig:model}~(a).

To address both the interpretability and scalability limitations of existing LBM works, we propose the Attribute-formed Language Bottleneck Model (ALBM). Compared with existing works, which organize the concept space in a class-shared manner by simply listing all concepts together, we propose to construct the Attribute-formed Class-specific Concept Space (ACCS). ACCS collects concept descriptions for each class under the guidance of a cross-category unified attribute set across categories, where \textit{attributes are general characteristics used to describe classes} (e.g., color, shape, texture), and \textit{concepts are specific descriptions based on attributes}, such as ``dark grey'' for the color of sooty albatross, as shown in Fig.~\ref{fig:model}~(b). By learning the concept classifier from ACCS, ALBM can address the limitations of scalability and interpretability in existing works. Specifically, ACCS defines the correspondence between classes and concepts during its construction, allowing our approach to predict categories solely based on class-specific essential concepts, thereby avoiding the spurious cue inference problem. As for scalability, attributes are more generalized than concepts and typically remain unchanged for unseen classes. Therefore, introducing novel classes only requires constructing their concept spaces based on the unified attribute set, leaving the concept spaces of seen classes unchanged. Moreover, since ACCS is built on the unified attribute set, correlations between concept spaces facilitate cross-class knowledge transfer. Thus, our approach can easily be generalized to novel classes.

\begin{figure*}[t]
	\centering
	{\includegraphics[width=\linewidth]{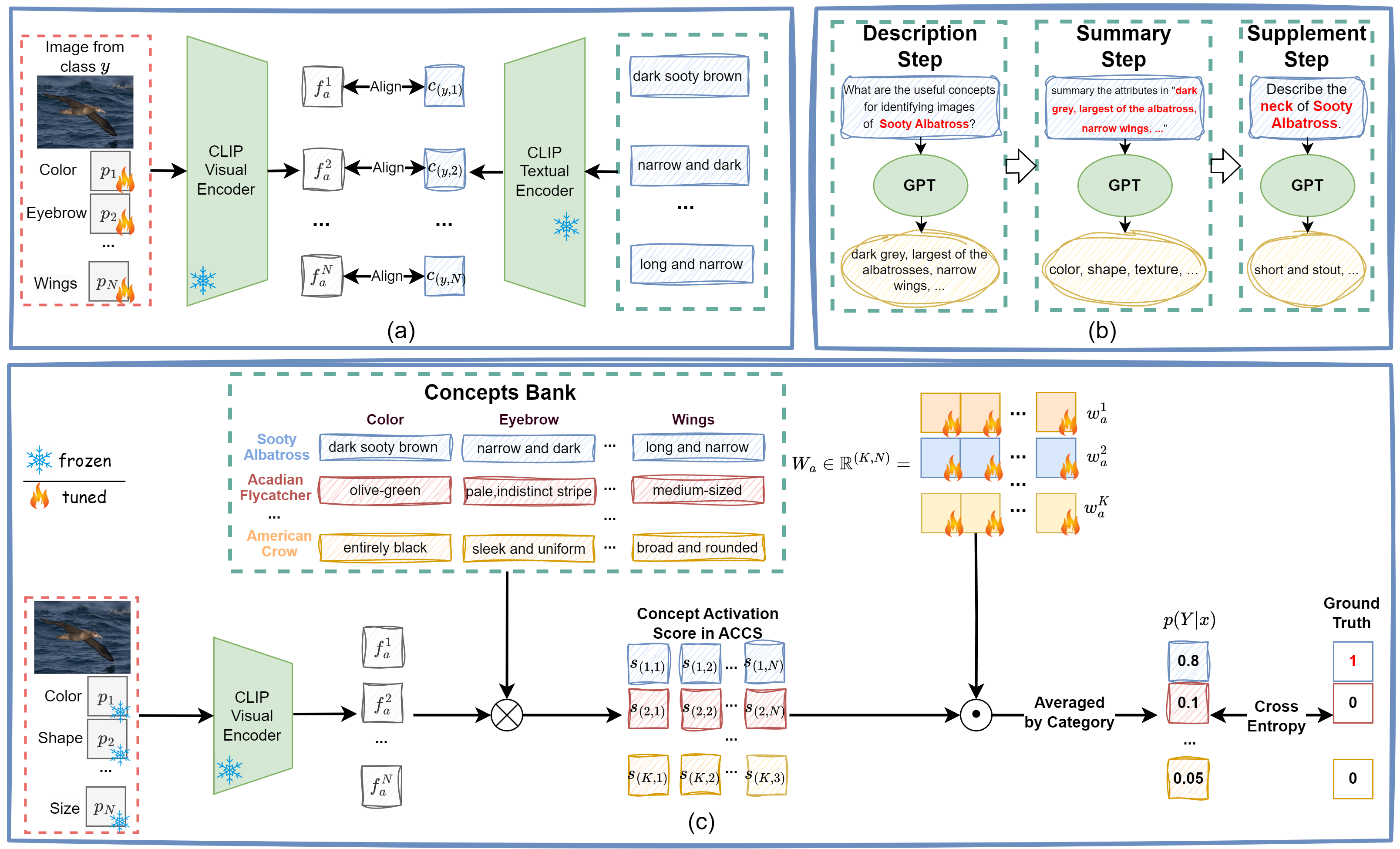}}
	\vspace{-0.7cm}
	\caption{(a) Illustration of Visual Attribute Prompt Learning (VAPL). VAPL trains visual prompts representing the semantics of each attribute by aligning the output feature of these prompts with the textual features of corresponding concepts. (b) Illustration of the Description, Summary, and Supplement (DSS) strategy. DSS first prompts the LLM to generate concepts for each class, then summarizes the corresponding attributes for each concept, and finally supplements missing attribute descriptions for each class. (c) The overall architecture of the Attribute-formed Language Bottleneck Model, where  $\otimes$ indicates matrix multiplication and $\odot$ indicates element-wise multiplication.}
	\label{fig:DSS}
	\vspace{-0.5cm}
\end{figure*}

Furthermore, the interpretability of existing LBMs is also limited by the quality of visual representations. LBMs require calculating activation scores of images on fine-grained features. However, the visual features extracted by the visual encoder of VLMs struggle to comprehensively capture the fine-grained characteristics of samples~\cite{DBLP:conf/iclr/GandelsmanES24}. To address this limitation, based on ACCS, we further propose Visual Attribute Prompt Learning (VAPL) to extract fine-grained visual features for each specific attribute. Specifically, different from conventional approaches~\cite{DBLP:conf/icml/RadfordKHRGASAM21, DBLP:conf/cvpr/YangPZJCY23} using the output feature of the [CLS] embedding as the overall representation of the image, we instead use the final output of the prompt of each attribute as the feature that represents the information of the input image on the corresponding attribute. However, unlike in the textual encoder, where attribute embeddings can be directly obtained by their text, what prompts can represent these attributes in the visual encoder is unknown. Therefore, to learn these attribute prompts, we align their output features with the concept descriptions of the corresponding attributes for the input samples. In this way, the learned attribute prompts can guide the visual encoder in capturing information about the input image on the corresponding attributes, which improves the accuracy of concept recognition and enhances overall explainable classification performance.

Moreover, to avoid labor-intensive and expensive manual concept collections, we propose to use LLM to automatically generate the class concepts. Currently, several existing works \cite{DBLP:conf/iccvw/ManiparambilVMM23, DBLP:conf/iclr/LiuRLZS024} have proposed generating class visual descriptions with unified attribute set based on chain-of-thought prompting \cite{DBLP:conf/nips/Wei0SBIXCLZ22} as follows: first asks LLM to list all attributes related to the target classes, and then use LLM to generate the descriptions of each class on these attributes. However, it is a difficult job for LLM to directly summarize the complete and precise attribute set that is required for identifying the target classes. Therefore, to generate high quality attribute-formed class-specific concept set, we further propose the Description, Summary, and Supplement (DSS) strategy as shown in Fig.~\ref{fig:DSS}~(b), which first prompts the LLM to generate representative concepts for each class, and then summarize the corresponding attributes for each concept and organize them into an attribute set. Finally, we use the LLM to supplement missing attribute descriptions for each class. By extracting attributes from class concepts rather than directly asking LLM for useful attributes, DSS can extract a more complete and precise attribute set.

We conduct extensive experiments on nine widely used fine-grained benchmarks, and the results demonstrate the effectiveness of our approach. The contributions of our work are mainly three-fold:

\begin{itemize}
	\item Analyze the limitations of existing LBMs in interpretability and scalability and further propose the Attribute-formed Language Bottleneck Model (ALBM) to address these limitations, by constructing the Attribute-formed Class-specific Concept Space to facilitate reasoning based on essential concepts of the corresponding class and cross-category knowledge transfer.
	\item Propose Visual Attribute Prompt Learning (VAPL) to extract visual features on each fine-grained attribute to improve the accuracy of concept recognition and further enhance the interpretability of LBM. 
	\item Propose the Description, Summary, and Supplement (DSS) mechanism to automatically generate the high-quality concept set with a complete and precise attribute set based on LLMs. 
\end{itemize}

\section{Related Works}
\noindent \textbf{Visual-Language Model.}
Existing researchers have found that the VLMs pre-trained by contrastive learning on large-scale internet data (e.g., CLIP \cite{DBLP:conf/icml/RadfordKHRGASAM21} and ALIGN \cite{DBLP:conf/icml/JiaYXCPPLSLD21}) have achieved great visual representation ability, resulting in remarkable performance in downstream tasks, image classification and image-text retrieval, and more. However, recent researchers found that as VLMs are pre-trained on unsupervised internet data, the visual feature extracted by VLMs may be mistakenly guided to align with text lacking visual information, resulting in insufficient semantics and interpretability of their visual representations \cite{DBLP:journals/tmlr/YunBPS23, DBLP:journals/corr/abs-2403-16442}.

\noindent \textbf{Training-free Language Bottleneck.}
Based on the concerns of insufficient interpretability and semantic richness of class name embedding, the existing researchers have proposed
Training-free Language Bottleneck \cite{DBLP:conf/iclr/MenonV23, DBLP:conf/iccv/PrattCLF23}, which proposes to use Large Language Model (LLM) \cite{DBLP:conf/nips/BrownMRSKDNSSAA20, DBLP:journals/corr/abs-2307-09288, DBLP:journals/corr/abs-2310-06825} to generate class descriptions in order to construct the concept set, and further achieves interpretable classification based on matching images with class concepts.

\noindent \textbf{Concept Bottleneck Model.}
Conventional deep image recognition models directly classify samples based on their extracted unexplainable image features \cite{DBLP:conf/cvpr/HeZRS16, DBLP:conf/iclr/DosovitskiyB0WZ21, DBLP:conf/icml/RadfordKHRGASAM21}. The unexplainability of this process makes it hard for developers to find and intervene the errors in these models. To improve the interpretability in image recognition, \cite{pmlr-v119-koh20a} proposed the Concept Bottleneck Model, which first extracts the concepts of the input sample. 
Recently, with growing interest in how VLMs make decisions,
\cite{DBLP:conf/cvpr/YangPZJCY23, DBLP:conf/iclr/OikarinenDNW23, DBLP:conf/iclr/YuksekgonulW023} proposed to achieve the explainable image recognition learning process for VLMs by integrating all class concepts into a unified concept space and learning a concept classifier based on this space. On this basis, \cite{DBLP:conf/iccv/0003WZDHLWSM23} further proposed a learning-to-search method to discover a much smaller concise set of concepts while maintaining the original classification performance. Moreover, \cite{shang2024incremental} proposed to mine missing concepts based on discovering the semantics of the learnable concept bottleneck residual. 
However, these existing LBMs learn concept classifiers in a class-shared concept space, leading to the spurious cue inference problem, and making them cannot transfer to unseen classes.

\noindent \textbf{Visual Prompt Learning.} Visual Prompt Learning is a kind of parameter-efficient tuning method for Vision Transformers \cite{DBLP:conf/iclr/DosovitskiyB0WZ21, DBLP:conf/iccv/LiuL00W0LG21}, which introduces additional learnable prompts to improve model performance on downstream tasks without fine-tuning the entire model. Generally, the prompts are introduced as the context of input embeddings \cite{DBLP:conf/eccv/JiaTCCBHL22} or the background of the input image \cite{bahng2022exploring}, thereby adjusting the model’s interpretation of the input image. Based on these approaches, Liu et al. \cite{liu2024multi} propose the Multi-modal Attribute Prompting (MAP) to capture visual details by aligning visual attribute prompts with class descriptions. However, the alignment process in MAP is promiscuous as the use of cross-attention and unstructured class descriptions without a unified attribute set. Therefore, its learned visual attribute prompts lack interpretable semantics. On the contrary, our approach aligns visual attribute prompts with concepts on specific attributes, such as ``color'' and ``texture'', which ensures that the learned prompts carry explicit and interpretable semantics.

\section{Method}

\subsection{Preliminaries}
\noindent \textbf{Revisiting CLIP.} CLIP \cite{DBLP:conf/icml/RadfordKHRGASAM21} consists of a visual encoder $\mathcal{V}$ and a language encoder $\mathcal{G}$ and bridges these two modalities by contrastive learning on Internet data. After pretraining, CLIP can be applied to downstream classification tasks in two paradigms, zero-shot classification and linear probe. Given the test sample $x$, the image encoder $\mathcal{V}$ first encodes $x$ into visual feature ${\bm{f}}=\mathcal{V}\left(x\right)$. For both zero-shot and linear probe classification, the prediction probability that the image
$x$ belongs to class $i$ is calculated as:
\begin{equation}
	p\left(y=i|x\right) = \frac{\exp\left({\rm sim} \left(\bm{f}, \bm{w}_i \right)/\tau\right)}{{\sum_{j=1}^{K} \exp\left({\rm sim} \left(\bm{f}, \bm{w}_j \right)/\tau\right)} },
\end{equation}
where $K$ is the number of classes, $\bm{w}_i$ is the classification weight of class $i$, and ${sim}\left(\cdot,\cdot  \right)$ is the cosine similarity.
The difference between zero-shot and linear probe classification is that for zero-shot classification, the classification weight $\bm{w}_z^i$ is obtained by extracting the text features of class name prompts (e.g., ``a photo of a $[\rm{class name}]_{i}$''), formally, $\bm{w}_z^i=\mathcal{G}\left(q_i\right)$, where $q_i$ is the class name prompt of the $i$-th class. Conversely, for linear probe classification, the classification weight $\bm{W}_l=[\bm{w}_l^1,\bm{w}_l^2,\ldots,\bm{w}_l^K]^T$ is a learnable parameter, which is optimized using the training dataset $\mathfrak{D}_{s}=\left\{ \left( x,y\right) \right\}$.

\noindent \textbf{Language Bottleneck Model.} 
LBMs \cite{DBLP:conf/iclr/YuksekgonulW023, DBLP:conf/cvpr/YangPZJCY23, DBLP:conf/iccv/0003WZDHLWSM23, shang2024incremental} achieve interpretable visual recognition by projecting the visual feature $\bm{f}$ directly onto the concept set ${\bm{C}} \in \mathbb{R}^{ N, d} $, and learn a linear concept classifier $\bm{W}_p=[\bm{w}_p^1,\bm{w}_p^2,\ldots,\bm{w}_p^K]^T \in \mathbb{R}^{K, N } $ to predict labels based on the concept scores, where $N$ is the number of concepts, and $d$ is the dimension of visual feature, formally,
\begin{equation}
	p\left(Y=y|x\right) = \frac{\exp\left(\bm{w}_p^y \cdot {\bm{C}} \cdot \bm{f}^T\right)}{{\sum_{i=1}^{K} \exp\left(\bm{w}_p^i \cdot {\bm{C}} \cdot \bm{f}^T\right)} }.
    \label{lbm-prediction}
\end{equation}

\begin{table*}[t]
	\begin{center}
		\setlength{\tabcolsep}{2.2mm}
		\begin{tabular}{cccccccccc}
			\toprule
			& Aircraft & CUB & DTD & Flowers102 & Food101 & OxfordPets & CIFAR-10 & CIFAR-100 & ImageNet\\
			\midrule
			CLIP-GPT & 22       & 7   & 8   & 18         & 16      & 7 & -  &  - &  17\\
			ALBM      & 23       & 37  & 33  & 26         & 29      & 12   & 11 & 21 & 55\\  \bottomrule
		\end{tabular}
		
	\end{center}
	\vspace{-0.3cm}
	\caption{The size of the attribute set on each dataset.}
	\vspace{-0.2cm}
	\label{attribute_number}
\end{table*}

\subsection{Attribute-formed language bottleneck model}
As discussed above, the LBMs face the spurious cue inference problem and cannot transfer to untrained novel classes. To improve the interpretability and transferability of LBM, we propose the Attribute-formed Language Bottleneck Model (ALBM), which classifies images in the Class-specific Concept Space (ACCS), the overall architecture of ALBM is shown in Fig.~\ref{fig:DSS}~(c). Specifically, our concept set ${\bm{\mathsfit{C}}} \in \mathbb{R}^{K,  N_a, d}$ is composed of the descriptions of all the attributes from a unified attribute set $\sA=\left\{a_j\right\}_{j=1}^{N_a}$ for all classes, where $N_a$ is the number of attributes, and the details of collecting $\sA$ are introduced in Section~\ref{DSS} and \ref{sec:DSS}.

By calculating the concept activation score $\mS={\tens{C}} \cdot \bm{f}^T \in \mathbb{R}^{K, N_a}$, we can obtain the representation of the input image $x$ in ACCS. Based on $\mS$, the linear concept classifier $\bm{W}_a \in \mathbb{R}^{K, N_a}$ can be learned to calculate the class prediction score, formally,
\begin{equation}
	\label{prediction}
	p\left(Y=j|x\right) = \frac{\exp\left(\bm{w}_a^j \cdot \vs_y^T\right)}{{\sum_{i=1}^{K} \exp\left(\bm{w}_a^i \cdot \vs_i^T \right)} },
\end{equation}
where $\bm{w}_a^i$ is the $i$-th column of $\bm{W}_a$, $\vs_i$ is the $i$-th column of $\mS$, which represents the concept activation score between $\bm{f}$ and the concepts of class $i$. Based on Eq.~\ref{prediction}, we can use cross-entropy loss $\mathcal{L}_w$ to optimize $\bm{W}_a$, formally,
\begin{equation}
	\mathcal{L}_w = \frac{1}{|\mathfrak{D}_{s}|}\sum_{x \in \mathfrak{D}_{s}}-\log \left(p\left(Y=y|x\right)\right).
\end{equation}
Notably, comparing Eq.~\ref{prediction} with Eq.~\ref{lbm-prediction}, we can see that different from existing LBMs classifying images based on class-shared concepts, our approach predicts the probability of a sample belonging to a class solely based on the class essential concepts but not concepts that are not causally related to the class. This helps us avoid the common problem of spurious cue inference in existing LBM approaches. 

Furthermore, since introducing novel classes will not change the concept space of base classes, and the attribute-formed concept spaces of different classes exhibit strong correlations guaranteed by sharing the unified attribute set, we can transfer the linear concept classifier $\bm{W}_a$ learned on base classes to unseen classes. Specifically, we first calculate the similarity between the target class and base classes, then weight the base class classifiers according to this similarity to integrate them as the novel class classifier, formally,
\begin{equation}
	\bm{w}_a^{K+j} = \sum_{i=1}^{K} \left(\frac{\exp\left(\bm{n}_i \cdot \bm{n}_{K+j}^T\right)}{{\sum_{l=1}^{K} \exp\left(\bm{n}_l \cdot \bm{n}_{K+j}^T \right)} } \cdot \bm{w}_a^i\right),
\end{equation}
where $\bm{w}_a^{K+j}$ is the attribute weight for the $j$-th novel classes, $\bm{n}_i$ is the class name feature of the $i$-th class. 
\subsection{Visual attribute prompt learning}	
Based on the motivation of extracting detailed visual features for each specific attribute, we propose Visual Attribute Prompt Learning (VAPL) as shown in Fig.~\ref{fig:DSS}~(a). VAPL uses a series of learnable visual prompts $\left \{ \vp_i \right \}_{j=1} ^{N_a}$ as the visual semantic embeddings for the attribute set $\left\{a_j\right\}_{j=1}^{N_a}$ and uses the output features of these prompts to represent the information of the input image on these attributes (e.g., color, texture, and shape), formally, 
\begin{equation}
	\left [ \vf_a^1, \vf_a^2, ..., \vf_a^N \right ] = \mathcal{V}\left(\left [ x, \vp_1, \vp_2, ..., \vp_N \right ] \right),
\end{equation}
where $\vf_a^i$ is the feature that represents information of $x$ on the $i$-th attribute. Notably, to prevent these visual attribute prompts from interfering with the feature extraction process, we masked the attention between these prompts and the attention from the image tokens to these prompts. Based on $\left [ \vf_a^1, \vf_a^2, ..., \vf_a^N \right ]$, the concept activation score $\mS$ can be recalculated as:
\begin{equation}
	s_{\left(i,j\right)} = {\rm sim} \left(\vf_a^j,\vc_{\left(i,j\right)}\right),
\end{equation}
where $s_{\left(i,j\right)}$ is the concept activation score of $x$ on the $j$-th concept of $i$-th class, and $\vc_{\left(i,j\right)}$ is the textual feature of the $j$-th concept of $i$-th class.

As the visual attribute prompts are initially undefined, we propose to learn these prompts by aligning each attribute feature $\vf_a^j$ with the corresponding concept descriptions $\vc_{\left(y,j\right)}$ based on the cross entropy loss, formally,
\begin{equation}
	\mathcal{L}_p = \frac{1}{N_a}\sum_{j=1}^{N_a}-\log \left(\frac{\exp\left(s_{\left(y,j\right)}\right)}{{\sum_{i=1}^{K} \exp\left( s_{\left(i,j\right)}\right)} }\right),
\end{equation}
where $y$ is the ground truth label of the input image.
By optimizing $\mathcal{L}_p$, the learned attribute prompts can guide visual encoder to capture information of the input image on the corresponding attributes.

\subsection{Description, summary, and supplement strategy}	
\label{DSS}
As discussed in Section~\ref{introduction}, we plan to use LLM to automatically generate the concept set to avoid labor-intensive and expensive manual concept collection. Based on concerns that it is difficult for LLMs to directly summarize the complete and precise attribute set for identifying the target classes, we propose the Description, Summary, and Supplement (DSS) strategy as shown in Fig.~\ref{fig:DSS}~(b), which first prompts the LLM to generate representative concepts for each class, and then summarize the attributes of these concepts. Finally, we use the LLM to supplement missing attribute values for each class. By summarizing attributes from freely generated class concepts, DSS can collect the attributes that are useful to identify categories in the target dataset as completely as possible. 

\noindent \textbf{Description step.}  
The first step of our DSS strategy is using LLM to freely generate concepts for each class by giving the class name $g_i$ and prompt $q_{des}$, formally,
\begin{equation}
	{\bm{c}}_i = {\rm LLM}\left( g_i, q_{des} \right),
\end{equation}
where $c_i$ is the original concept set of class $i$.
As this step is consistent with the implementation of existing language bottleneck approaches \cite{DBLP:conf/iclr/MenonV23, DBLP:conf/cvpr/YangPZJCY23}, we can directly use their collected concept sets to avoid redundant calculations.

\noindent \textbf{Summary step.}  
After getting the concepts of classes, our goal is to summarize the attributes corresponding to these concepts by giving LLM all the generated concepts $\left\{{\bm{c}}_i\right\}_{i=1}^K$ in the description step and prompt  $q_{sum}$, formally,
\begin{equation}
	\mA = {\rm LLM}\left( \left\{{\bm{c}_i}\right\}_{i=1}^K, q_{sum} \right).
\end{equation}
The detail of this summary step can be seen in Appendix~\ref{sec:DSS}.

\noindent \textbf{Supplement step.} Due to the lack of a unified attribute set in the description step, some categories have missing concept descriptions on several attributes. Therefore, we further utilize LLM to supply these missing concepts with the prompt $p_{sup}$, formally,
\begin{equation}
	{\bm{c}}_i^j = {\rm LLM}\left(g_i, a_j, q_{vis} \right),
\end{equation}
where ${\bm{c}}_i^j$ is the missing concept of class $i$ for attribute $a_j$.

By implementing the above three steps, a concept set with rich attributes can be obtained. Based on it, we can achieve better interpretability and transferability compared with existing LBMs.

\section{Experiments}
\subsection{Experimental setup}

\noindent \textbf{Dataset.} Following existing Training-free Language Bottleneck approaches \cite{DBLP:conf/iccv/0003WZDHLWSM23, DBLP:conf/iccv/PrattCLF23} and Language Bottleneck Model \cite{DBLP:conf/cvpr/YangPZJCY23}, we conduct experiments on 8 widely used few-shot benchmarks: including Aircraft~\cite{DBLP:journals/corr/MajiRKBV13} for airplane recognition, CUB~\cite{wah2011caltech} for fine-grained bird categories, DTD~\cite{DBLP:conf/cvpr/CimpoiMKMV14} for textures, Flowers102~\cite{DBLP:conf/icvgip/NilsbackZ08} for flowers, Food101~\cite{DBLP:conf/eccv/BossardGG14} for food, OxfordPets~\cite{DBLP:conf/cvpr/ParkhiVZJ12} for common animals, CIFAR-10, CIFAR-100~\cite{krizhevsky2009learning} and ImageNet~\cite{DBLP:journals/ijcv/RussakovskyDSKS15} for common objects.

\noindent \textbf{Compared methods.} We compare our approach with black-box CLIP model and the recent CLIP-based explainable image recognition approaches, including: (1) The original Zero-shot CLIP (ZS-CLIP) and Linear Probe CLIP (LP-CLIP)~\cite{DBLP:conf/icml/RadfordKHRGASAM21}; (2) the Training-free Language Bottleneck approaches: Visual Description CLIP (VDCLIP)~\cite{DBLP:conf/iclr/MenonV23}, CuPL~\cite{DBLP:conf/iccv/PrattCLF23}, and CLIP-GPT~\cite{DBLP:conf/iccvw/ManiparambilVMM23}; (3) the Language Bottleneck Models: LaBo~\cite{DBLP:conf/cvpr/YangPZJCY23} and Concise Language Bottleneck Model (CLBM)~\citep{DBLP:conf/iccv/0003WZDHLWSM23}.

\noindent \textbf{Implementation details.} In all experiments, the pre-trained ViT-L/14 CLIP model~\cite{DBLP:conf/icml/RadfordKHRGASAM21} is utilized as the visual and textual feature extractors for our approach and all comparison approaches. In addition, GPT-4o is utilized as the LLM in DSS. To avoid redundant calculations, we use the concept sets collected by \cite{DBLP:conf/cvpr/YangPZJCY23} as the result of the description step of DSS. As for the training process, We first train the visual attribute prompts using $\mathcal{L}_p$ with a learning rate of 0.0035, a batch size of 64, and running for 5 epochs. Next, we train 
$\mW_a$ using $\mathcal{L}_w$ with a learning rate of 0.0006, a batch size of 64, and running for 1000 epochs. SGD optimizer is used for all training processes. 

\begin{table}[t]

	\begin{center}
		\begin{tabularx}{\linewidth}{XX}
			\toprule
			\multicolumn{1}{c}{CLIP-GPT}                                                   & \multicolumn{1}{c}{ALBM}                                                         \\ \midrule
			size, length, fur texture,   fur color, eye color, ear shape, distinctive features & fur, size, breed, appearance, body, color, snout, head, legs,   tail, eyes, ears \\ \bottomrule
		\end{tabularx}
	\end{center}
    \vspace{-0.3cm}
	\caption{Comparison of the collected attributes on OxfordPets.}
	\label{attribute_case}
    \vspace{-0.3cm}
\end{table}
\subsection{Quality of the constructed attribute set}
We first validate the quality of the attribute set constructed by DSS. As shown in Tab.~\ref{attribute_number}, in general, compared with the existing attribute-formed concept collection approach CLIP-GPT~\cite{DBLP:conf/iccvw/ManiparambilVMM23}, our approach collected more attributes.
From Tab.~\ref{attribute_case}, we can see that our approach can list some important attributes which are useful for recognizing target classes that CLIP-GPT ignores (e.g., ``snout'', ``leg'', and ``tail'' for OxfordPets). This is because directly summarizing the attribute set, as CLIP-GPT does, is a hard mission for GPT. On the contrary, our approach summarizes attributes from the freely generated class concepts, which can extract a more complete and precise attribute set. 
\subsection{Performance comparison}

\begin{table*}[t]

		\resizebox{\linewidth}{!}{
        \setlength{\tabcolsep}{2mm}
        \begin{tabular}{ccccccccccc}
			\toprule
			&     Approach     & Aircraft & CUB           & DTD           & Flowers102    & Food101       & OxfordPets  & CIFAR-10 & CIFAR-100 & ImageNet \\ \midrule
			Unexplainable                        & ZS-CLIP~\cite{DBLP:conf/icml/RadfordKHRGASAM21} & 32.6     & 63.4          & 53.2          & 79.3          & 91.0          & 93.6        & 86.0 & 55.6  & 71.4\\ \midrule
			\multirow{5}{*}{\begin{tabular}[c]{@{}c@{}}Training-free\\ Language\\ Bottleneck\end{tabular}} & VDCLIP~\cite{DBLP:conf/iclr/MenonV23}   & -        & 3.9           & 18.6          & -             & 15.2           & 11.9 & - & -     &   23.4   \\
			& CuPL~\cite{DBLP:conf/iccv/PrattCLF23}     & \textbf{19.9}     & -             & 37.2          & -             & 66.3          & 33.9 & 75.2 & 40.6      &   59.2  \\
			& CLIP-GPT~\cite{DBLP:conf/iccvw/ManiparambilVMM23} & 13.4     & 11.4           & 40.0          & 11.7          & 48.4          & 31.9  & - & -     & 44.3  \\
			& LaBo*~\cite{DBLP:conf/cvpr/YangPZJCY23}     & 15.7     & 16.2          & 37.9          & 34.2          & 52.2          & - & 64.0 & 31.1      &    37.8   \\
			& ALBM* (ours)    & 18.0     & \textbf{25.0} & \textbf{48.5} & \textbf{54.9} & \textbf{75.4} & \textbf{35.9} & \textbf{83.1} & \textbf{43.1}& \textbf{64.6}\\ \bottomrule
		\end{tabular}}
\vspace{-0.2cm}
	\caption{Comparison with zero-shot CLIP and training-free language bottlenecks in the zero-shot setting. * denote zero-shot predictions based on their collected concept sets, while ``-'' indicates that the original approaches didn't collect a concept set for the dataset.}
    \vspace{-0.1cm}
	\label{ZSL}
\end{table*}

\begin{table*}[t]
	
	\begin{center}

\resizebox{\linewidth}{!}{\setlength{\tabcolsep}{0.7mm}\begin{tabular}{cccccccccccccccccccc}
\toprule
                                                     & \multirow{2}{*}{Approach} & \multicolumn{2}{c}{Aircraft}                & \multicolumn{2}{c}{CUB}                     & \multicolumn{2}{c}{DTD}                     & \multicolumn{2}{c}{Flowers102}              & \multicolumn{2}{c}{Food101}                 & \multicolumn{2}{c}{OxfordPets}              & \multicolumn{2}{c}{CIFAR-10}                 & \multicolumn{2}{c}{CIFAR-100}         &    \multicolumn{2}{c}{ImageNet}   \\ \cmidrule{3-20}
                                                     &                           & Base                 & Novel                & Base                 & Novel                & Base                 & Novel                & Base                 & Novel                & Base                 & Novel                & Base                 & Novel                & Base                 & Novel                & Base                 & Novel  & Base                 & Novel              \\ \midrule
\multirow{2}{*}{Unexplainable}                       & ZS-CLIP~\citep{DBLP:conf/icml/RadfordKHRGASAM21}                  & 37.2                 & 44.5                 & 69.9                 & 60.1                 & 61.2                 & 71.4                 & 83.2                 & 82.7                 & 93.7                 & 94.9                 & 95.1                 & 98.2                 & 91.1                 & 93.7                 & 66.9                 & 60.2               & 77.2& 72.3 \\
                                                     & LP-CLIP~\citep{DBLP:conf/icml/RadfordKHRGASAM21}                   & 50.9 & - & 86.4 & - & 80.7 & - & 98.6 & - & 91.6 & - & 93.3 & - & 91.1 & - & 71.3 & - & 78.5& -\\ \midrule
\multirow{3}{*}{\begin{tabular}[c]{@{}c@{}}Training-free\\ Language\\ Bottleneck\end{tabular}} & VD-CLIP~\citep{DBLP:conf/iclr/MenonV23}                   & -                    & -                    & 7.0                  & 5.5                  & 31.0                 & 21.4                 & -                    & -                    & 19.9                 & 18.0                 & 14.8                 & 22.5                 & -                    & -                    & -                    & -                   & 20.7& 34.0\\
                                                     & CuPL~\citep{DBLP:conf/iccv/PrattCLF23}                      & 22.7                 & 30.5                 & -                    & -                    & 51.2                 & 43.8                 & -                    & -                    & 71.6                 & 78.3                 & 42.8                 & 49.4                 & 88.5                 & 92.8                 & 49.4                 & 49.7                 & 59.8 & 66.8\\
                                                     & CLIP-GPT~\citep{DBLP:conf/iccvw/ManiparambilVMM23}                  & 14.2                 & 18.4                 & 18.6                 & 15.6                 & 52.0                 & 49.6                 & 11.0                 & 16.5                 & 60.8                 & 57.5                 & 46.0                 & 46.1                 & -                    & -                    & -                    & -                    & 70.0& 22.1\\ \midrule
\multirow{2}{*}{\begin{tabular}[c]{@{}c@{}}Language Bott-\\leneck Model\end{tabular}}   & LaBo~\citep{DBLP:conf/cvpr/YangPZJCY23}                      & \textbf{42.9}        & -                    & 76.9                 & -                    & 77.0                 & -                    & 87.6                 & -                    & \textbf{90.8}        & -                    & -                    & -                    & 89.6        & -                    &   55.6                   &     -                 & 71.7& -\\
& CLBM~\citep{DBLP:conf/iccv/0003WZDHLWSM23}                   & -                    & -                    & 67.4                 & -                    & -                    & -                    & 52.0                 & -                    & -                    & -                    & 60.0                 & -                    & -                    & -                    & 51.4                 & -                    & -& -\\ \midrule
Base-to-Novel                                        & ALBM (ours)               & 38.7                 & \textbf{33.0}        & \textbf{91.9}        & \textbf{27.8}        & \textbf{78.6}        & \textbf{60.5}        & \textbf{91.7}        & \textbf{32.4}        & 88.5                 & \textbf{86.8}        & \textbf{79.2}        & \textbf{61.1}        & \textbf{90.8}                 & \textbf{93.6}        & \textbf{59.3}        & \textbf{55.1}       & \textbf{75.0}& \textbf{73.9}\\ \bottomrule
\end{tabular}}
\vspace{-0.2cm}
	\end{center}
	\caption{Comparison with unexplainable CLIP, Training-free Language Bottlenecks, and Language Bottleneck Models on the base-to-novel setting, where Training-free Language Bottlenecks are zero-shot learning approaches, Language Bottleneck Models are trained on base classes, and ``-'' indicates that the original approaches didn't collect the concept set for the dataset or unavailable for novel classes.}
    \vspace{-0.3cm}
	\label{B2N}
\end{table*}

We conduct comprehensive experiments on the zero-shot setting, and base-to-novel setting to validate the interpretability, scalability, and performance of our approach, where in the zero-shot setting, classification is directly according to the average matching scores between the image feature and the corresponding concepts of each category, and in base-to-novel setting, each dataset is divided into subsets of base and novel classes, and only base classes provide 16 images of each class for training. Notably, to fairly compare the performance of explainable image recognition, we remove the class name carried in the concept descriptions for VDCLIP, CuPL, and CLIP-GPT.

\noindent \textbf{Zero-shot performance.} 
To validate the performance of our collected attribute-formed concept sets, we conduct comparison analysis on zero-shot setting.
As shown in Tab.~\ref{ZSL}, compared with existing Training-free Language Bottlenecks, our approach shows significant performance improvement on eight of the nine datasets, achieving 2.0\% to 20.7\% performance improvements compared with the best results of existing approaches, except on the Aircraft dataset, where there is a slight performance decrease of 1.9\% compared to the existing best result. Specifically, compared to CLIP-GPT, which is also built on a unified attribute set, our approach still demonstrates significant advantages. These experimental results show that our DSS strategy, by constructing a more comprehensive unified attribute set and collecting class concepts based on it, can gather a more complete and systematic semantic knowledge base of classes, thereby enhancing the effectiveness of interpretable image classification. However, there remains a substantial performance gap compared to unexplainable class-name-based image classification, highlighting the importance of learnable language bottleneck models.

\noindent \textbf{Base-to-novel performance.}
To validate the scalability of our proposed approach, we conduct comparison analysis on base-to-novel generalization setting, as shown in Tab.~\ref{B2N}. Specifically, compared with unexplainable CLIP, all interpretable approaches show lower performances. This is mainly caused by the insufficient interpretability of the visual features extracted by CLIP, as CLIP learns from noisy samples (the paired text may not clearly describe the visual information of the images) in the pretraining process. This phenomenon is consistent with findings in existing research~\cite{DBLP:journals/tmlr/YunBPS23, DBLP:journals/corr/abs-2403-16442}.

Compared with existing training-free language bottlenecks, we achieve significant performance superiority across all base and novel classes, achieving 1.0\% to 80.7\% improvement on base classes and 0.6\% to 15.9\% improvement on novel classes. The significant performance boost observed on base classes underscores the importance of the learnable LBM methodology, while the remarkable improvements on novel classes validate the effectiveness of our proposed approach in terms of cross-category scalability, which is consistent with our motivation. 

Compared with existing Language Bottleneck Models, our approach can also achieve improved performance on 7 out of 9 datasets (improvements ranging from 1.3\% to 19.2\%), and slight performance drops from 2.3\% to 4.2\% on the rest two datasets compared to the existing LBMs. Notably, as discussed above, existing LBMs learn the concept classifier in the class-shared concept space, which may recognize classes based on spurious cues. These spurious cues enable the classifier to gain performance from inexplicable representations, thereby weakening the interpretability ``bottleneck''. On the contrary, ALBM is constrained to identify categories solely through the essential features of each category. Therefore, considering that our approach is more reliable in interpretability and can be transferred to novel classes, achieving comparable or even superior performance on base classes relative to existing LBM methods represents a noteworthy outcome.

\subsection{Analysis}
\begin{table}[t]

\begin{minipage}{0.35\linewidth} \caption{Ablation study on the proposed components. Averages across the nine datasets are reported.} 
\label{ablation}
\end{minipage}
\begin{minipage}{0.65\linewidth}
\centering
\begin{tabular}{cccc}
\toprule
$\mathcal{L}_w$ & $\mathcal{L}_p$ & Base & Novel \\ \midrule
\xmark    & \xmark    & 54.2 & 55.2  \\
\checkmark & \xmark    &  74.0    &    57.5   \\
\checkmark    & \checkmark    & \textbf{77.2} & \textbf{58.2} \\ \bottomrule
\end{tabular}
\end{minipage}
\vspace{-0.5cm}
\end{table}

\begin{figure*}[t!]
	\centering
	{\includegraphics[width=\linewidth]{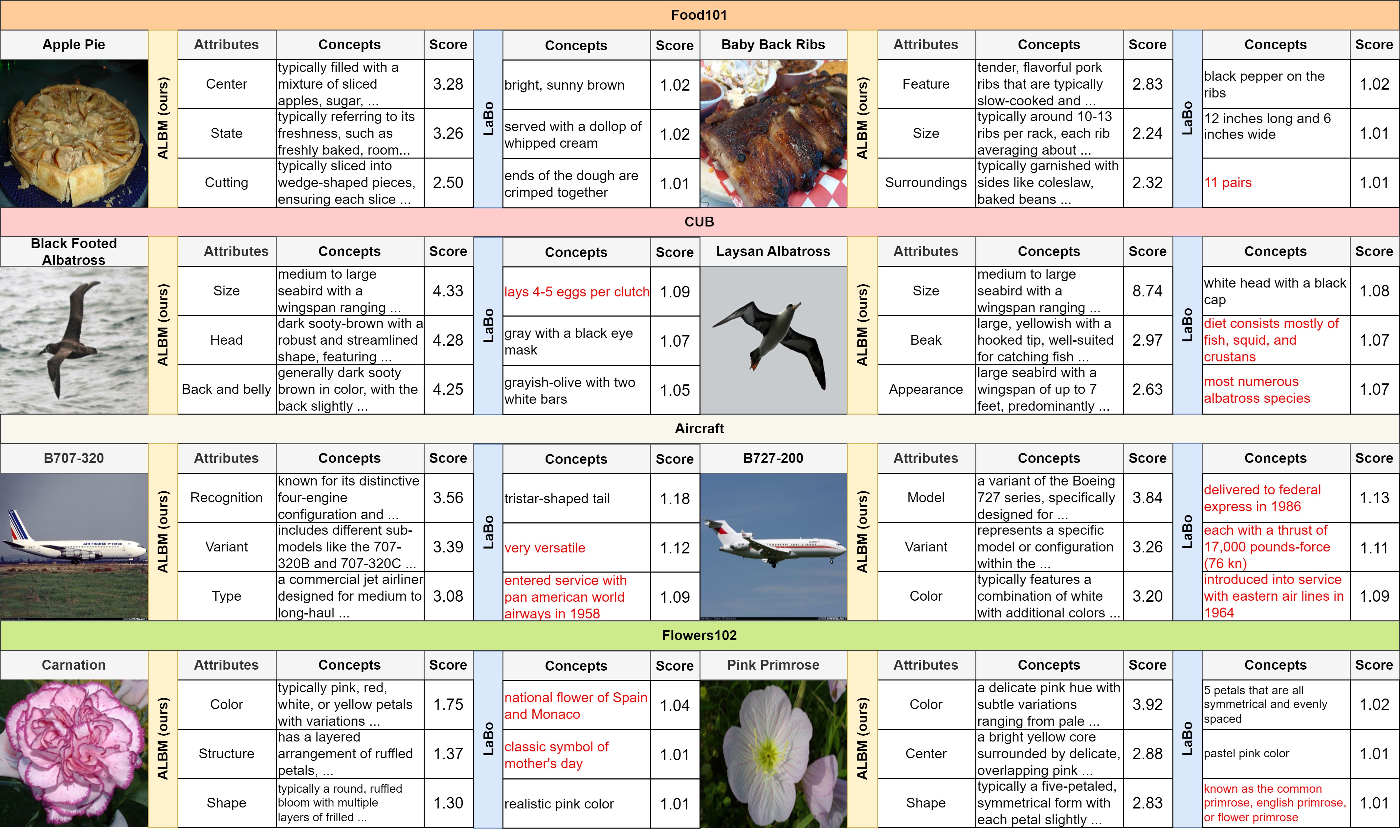}}
	\caption{Case study of bottlenecks constructed by ALBM and LaBo, where red texts indicate spurious cues, scores indicate concept activations. The top three highest-weighted concepts for each category are shown. Categories and datasets are selected randomly.}
	\label{fig:visual}
	\vspace{-0.35cm}
\end{figure*}

\noindent \textbf{Ablation study.}
To further demonstrate the effectiveness of the proposed VAPL and ALBM, we conduct
ablation studies and show the corresponding results in Tab.~\ref{ablation}. Notably, removing $\mathcal{L}_w$ represents the results of zero-shot classification. Comparing the first and second lines, it is clear that training the concept classifier achieves 19.8\% and 2.3\% performance improvements on base and novel classes, respectively. The performance improvement on base classes demonstrates the importance of learning a concept classifier. Additionally, the improvement on novel classes indicates that our attribute-formed paradigm effectively facilitates cross-category transfer, enhancing the scalability of LBM. Moreover, comparing the second and third lines, we can find that VAPL achieves 3.2\% and 0.7\% performance improvements on base and novel classes, respectively. These results show that the visual attribute prompts can better capture fine-grained features of images, thereby enhancing explainable image recognition. The performance improvement on novel classes further demonstrates the generalizability of the visual attribute prompts, i.e., even for samples from untrained classes, VAPL can also help extract their fine-grained features.

\noindent \textbf{Case study of interpretability.}
To verify the interpretability of our approach, we compare bottlenecks constructed by our approach and existing LBM approach LaBo as shown in Fig.~\ref{fig:visual}. It is clear that learning the concept classifier from a class-shared concept space (as the implementation of existing LBMs) constructs a concept bottleneck that may be based on spurious cues (indicated by red text in the figure), which limits the interpretability of existing LBM approaches. In contrast, our approach constructs a unified visual attribute set, creates an attribute-formed class-specific concept space based on this set, and learns the concept classifier within this space. In this way, our approach recognizes the class solely based on class-specific essential concepts, avoiding the problem of spurious cue inference and achieving more reliable interpretability.
\section{Conclusion}
In this work, we analyze the limitations of interpretability and scalability in existing LBMs, which arise from the risk of inference based on spurious cues and the expansion of the concept space when adding new classes. To address these limitations, we propose the Attribute-formed Language Bottleneck Model.
By building ACCS, our approach predicts labels solely based on class-corresponding essential concepts to avoid the spurious cue inference problem. Additionally, our approach can easily generalize to novel classes based on the cross-category consistent attribute set.
Moreover, we further propose VAPL to extract visual features on each fine-grained attribute to improve the accuracy of concept recognition and further enhance LBM performance. 
Furthermore, by employing the DSS strategy, we automate the creation of concept set with a high-quality unified attribute set by summarizing general attributes from freely generated class concepts. The experimental results on nine widely used benchmarks validate the effectiveness of both our proposed ALBM approach and the collected concept sets. Future work will focus on improving the performance of the explainable Visual-Language Model and improving the interpretability of the visual feature extracted by VLM based on our ALBM.

\section*{Acknowledgments}
This work was supported by the National Natural Science Foundation of China (No. 62106204, 62172075), the Natural Science Foundation of Sichuan (No. 2025YFHZ0124), the Frontier Cross Innovation Team Project of Southwest Jiaotong University (YH1500112432297), the Natural Science Foundation of Xinjiang Uygur Autonomous (No. 2024D01A14), and the CSC scholarship.

{
    \small
    \bibliographystyle{ieeenat_fullname}
    \bibliography{main}

\begin{thebibliography}{35}
\providecommand{\natexlab}[1]{#1}
\providecommand{\url}[1]{\texttt{#1}}
\expandafter\ifx\csname urlstyle\endcsname\relax
  \providecommand{\doi}[1]{doi: #1}\else
  \providecommand{\doi}{doi: \begingroup \urlstyle{rm}\Url}\fi

\bibitem[Bahng et~al.(2022)Bahng, Jahanian, Sankaranarayanan, and
  Isola]{bahng2022exploring}
Hyojin Bahng, Ali Jahanian, Swami Sankaranarayanan, and Phillip Isola.
\newblock Exploring visual prompts for adapting large-scale models.
\newblock \emph{arXiv preprint arXiv:2203.17274}, 2022.

\bibitem[Bossard et~al.(2014)Bossard, Guillaumin, and
  Gool]{DBLP:conf/eccv/BossardGG14}
Lukas Bossard, Matthieu Guillaumin, and Luc~Van Gool.
\newblock Food-101 - mining discriminative components with random forests.
\newblock In \emph{Computer Vision - {ECCV} 2014 - 13th European Conference,
  Zurich, Switzerland, September 6-12, 2014, Proceedings, Part {VI}}, pages
  446--461. Springer, 2014.

\bibitem[Brown et~al.(2020)Brown, Mann, Ryder, Subbiah, Kaplan, Dhariwal,
  Neelakantan, Shyam, Sastry, Askell, Agarwal, Herbert{-}Voss, Krueger,
  Henighan, Child, Ramesh, Ziegler, Wu, Winter, Hesse, Chen, Sigler, Litwin,
  Gray, Chess, Clark, Berner, McCandlish, Radford, Sutskever, and
  Amodei]{DBLP:conf/nips/BrownMRSKDNSSAA20}
Tom~B. Brown, Benjamin Mann, Nick Ryder, Melanie Subbiah, Jared Kaplan,
  Prafulla Dhariwal, Arvind Neelakantan, Pranav Shyam, Girish Sastry, Amanda
  Askell, Sandhini Agarwal, Ariel Herbert{-}Voss, Gretchen Krueger, Tom
  Henighan, Rewon Child, Aditya Ramesh, Daniel~M. Ziegler, Jeffrey Wu, Clemens
  Winter, Christopher Hesse, Mark Chen, Eric Sigler, Mateusz Litwin, Scott
  Gray, Benjamin Chess, Jack Clark, Christopher Berner, Sam McCandlish, Alec
  Radford, Ilya Sutskever, and Dario Amodei.
\newblock Language models are few-shot learners.
\newblock In \emph{Advances in Neural Information Processing Systems 33: Annual
  Conference on Neural Information Processing Systems 2020, NeurIPS 2020,
  December 6-12, 2020, virtual}, 2020.

\bibitem[Cimpoi et~al.(2014)Cimpoi, Maji, Kokkinos, Mohamed, and
  Vedaldi]{DBLP:conf/cvpr/CimpoiMKMV14}
Mircea Cimpoi, Subhransu Maji, Iasonas Kokkinos, Sammy Mohamed, and Andrea
  Vedaldi.
\newblock Describing textures in the wild.
\newblock In \emph{2014 {IEEE} Conference on Computer Vision and Pattern
  Recognition, {CVPR} 2014, Columbus, OH, USA, June 23-28, 2014}, pages
  3606--3613. {IEEE} Computer Society, 2014.

\bibitem[Dosovitskiy et~al.(2021)Dosovitskiy, Beyer, Kolesnikov, Weissenborn,
  Zhai, Unterthiner, Dehghani, Minderer, Heigold, Gelly, Uszkoreit, and
  Houlsby]{DBLP:conf/iclr/DosovitskiyB0WZ21}
Alexey Dosovitskiy, Lucas Beyer, Alexander Kolesnikov, Dirk Weissenborn,
  Xiaohua Zhai, Thomas Unterthiner, Mostafa Dehghani, Matthias Minderer, Georg
  Heigold, Sylvain Gelly, Jakob Uszkoreit, and Neil Houlsby.
\newblock An image is worth 16x16 words: Transformers for image recognition at
  scale.
\newblock In \emph{9th International Conference on Learning Representations,
  {ICLR} 2021, Virtual Event, Austria, May 3-7, 2021}. OpenReview.net, 2021.

\bibitem[Esfandiarpoor et~al.(2024)Esfandiarpoor, Menghini, and
  Bach]{DBLP:journals/corr/abs-2403-16442}
Reza Esfandiarpoor, Cristina Menghini, and Stephen~H. Bach.
\newblock If {CLIP} could talk: Understanding vision-language model
  representations through their preferred concept descriptions.
\newblock \emph{CoRR}, abs/2403.16442, 2024.

\bibitem[Gandelsman et~al.(2024)Gandelsman, Efros, and
  Steinhardt]{DBLP:conf/iclr/GandelsmanES24}
Yossi Gandelsman, Alexei~A. Efros, and Jacob Steinhardt.
\newblock Interpreting clip's image representation via text-based
  decomposition.
\newblock In \emph{The Twelfth International Conference on Learning
  Representations, {ICLR} 2024, Vienna, Austria, May 7-11, 2024}.
  OpenReview.net, 2024.

\bibitem[He et~al.(2016)He, Zhang, Ren, and Sun]{DBLP:conf/cvpr/HeZRS16}
Kaiming He, Xiangyu Zhang, Shaoqing Ren, and Jian Sun.
\newblock Deep residual learning for image recognition.
\newblock In \emph{2016 {IEEE} Conference on Computer Vision and Pattern
  Recognition, {CVPR} 2016, Las Vegas, NV, USA, June 27-30, 2016}, pages
  770--778. {IEEE} Computer Society, 2016.

\bibitem[Jia et~al.(2021)Jia, Yang, Xia, Chen, Parekh, Pham, Le, Sung, Li, and
  Duerig]{DBLP:conf/icml/JiaYXCPPLSLD21}
Chao Jia, Yinfei Yang, Ye Xia, Yi{-}Ting Chen, Zarana Parekh, Hieu Pham,
  Quoc~V. Le, Yun{-}Hsuan Sung, Zhen Li, and Tom Duerig.
\newblock Scaling up visual and vision-language representation learning with
  noisy text supervision.
\newblock In \emph{Proceedings of the 38th International Conference on Machine
  Learning, {ICML} 2021, 18-24 July 2021, Virtual Event}, pages 4904--4916.
  {PMLR}, 2021.

\bibitem[Jia et~al.(2022)Jia, Tang, Chen, Cardie, Belongie, Hariharan, and
  Lim]{DBLP:conf/eccv/JiaTCCBHL22}
Menglin Jia, Luming Tang, Bor{-}Chun Chen, Claire Cardie, Serge~J. Belongie,
  Bharath Hariharan, and Ser{-}Nam Lim.
\newblock Visual prompt tuning.
\newblock In \emph{Computer Vision - {ECCV} 2022 - 17th European Conference,
  Tel Aviv, Israel, October 23-27, 2022, Proceedings, Part {XXXIII}}, pages
  709--727. Springer, 2022.

\bibitem[Jiang et~al.(2023)Jiang, Sablayrolles, Mensch, Bamford, Chaplot,
  de~Las~Casas, Bressand, Lengyel, Lample, Saulnier, Lavaud, Lachaux, Stock,
  Scao, Lavril, Wang, Lacroix, and Sayed]{DBLP:journals/corr/abs-2310-06825}
Albert~Q. Jiang, Alexandre Sablayrolles, Arthur Mensch, Chris Bamford,
  Devendra~Singh Chaplot, Diego de Las~Casas, Florian Bressand, Gianna Lengyel,
  Guillaume Lample, Lucile Saulnier, L{\'{e}}lio~Renard Lavaud, Marie{-}Anne
  Lachaux, Pierre Stock, Teven~Le Scao, Thibaut Lavril, Thomas Wang,
  Timoth{\'{e}}e Lacroix, and William~El Sayed.
\newblock Mistral 7b.
\newblock \emph{CoRR}, abs/2310.06825, 2023.

\bibitem[Kim et~al.(2023)Kim, Kim, Choi, and Kim]{kim2023concept}
Injae Kim, Jongha Kim, Joonmyung Choi, and Hyunwoo~J Kim.
\newblock Concept bottleneck with visual concept filtering for explainable
  medical image classification.
\newblock In \emph{International Conference on Medical Image Computing and
  Computer-Assisted Intervention}, pages 225--233. Springer, 2023.

\bibitem[Koh et~al.(2020)Koh, Nguyen, Tang, Mussmann, Pierson, Kim, and
  Liang]{pmlr-v119-koh20a}
Pang~Wei Koh, Thao Nguyen, Yew~Siang Tang, Stephen Mussmann, Emma Pierson, Been
  Kim, and Percy Liang.
\newblock Concept bottleneck models.
\newblock In \emph{Proceedings of the 37th International Conference on Machine
  Learning}, pages 5338--5348. PMLR, 2020.

\bibitem[Krizhevsky et~al.(2009)Krizhevsky, Hinton,
  et~al.]{krizhevsky2009learning}
Alex Krizhevsky, Geoffrey Hinton, et~al.
\newblock Learning multiple layers of features from tiny images.
\newblock 2009.

\bibitem[Liu et~al.(2024{\natexlab{a}})Liu, Roy, Li, Zhong, Sebe, and
  Ricci]{DBLP:conf/iclr/LiuRLZS024}
Mingxuan Liu, Subhankar Roy, Wenjing Li, Zhun Zhong, Nicu Sebe, and Elisa
  Ricci.
\newblock Democratizing fine-grained visual recognition with large language
  models.
\newblock In \emph{The Twelfth International Conference on Learning
  Representations, {ICLR} 2024, Vienna, Austria, May 7-11, 2024}.
  OpenReview.net, 2024{\natexlab{a}}.

\bibitem[Liu et~al.(2024{\natexlab{b}})Liu, Wu, Yang, Zhou, and
  Zhang]{liu2024multi}
Xin Liu, Jiamin Wu, Wenfei Yang, Xu Zhou, and Tianzhu Zhang.
\newblock Multi-modal attribute prompting for vision-language models.
\newblock \emph{IEEE Transactions on Circuits and Systems for Video
  Technology}, 2024{\natexlab{b}}.

\bibitem[Liu et~al.(2021)Liu, Lin, Cao, Hu, Wei, Zhang, Lin, and
  Guo]{DBLP:conf/iccv/LiuL00W0LG21}
Ze Liu, Yutong Lin, Yue Cao, Han Hu, Yixuan Wei, Zheng Zhang, Stephen Lin, and
  Baining Guo.
\newblock Swin transformer: Hierarchical vision transformer using shifted
  windows.
\newblock In \emph{2021 {IEEE/CVF} International Conference on Computer Vision,
  {ICCV} 2021, Montreal, QC, Canada, October 10-17, 2021}, pages 9992--10002.
  {IEEE}, 2021.

\bibitem[Maji et~al.(2013)Maji, Rahtu, Kannala, Blaschko, and
  Vedaldi]{DBLP:journals/corr/MajiRKBV13}
Subhransu Maji, Esa Rahtu, Juho Kannala, Matthew~B. Blaschko, and Andrea
  Vedaldi.
\newblock Fine-grained visual classification of aircraft.
\newblock \emph{CoRR}, abs/1306.5151, 2013.

\bibitem[Maniparambil et~al.(2023)Maniparambil, Vorster, Molloy, Murphy,
  McGuinness, and O'Connor]{DBLP:conf/iccvw/ManiparambilVMM23}
Mayug Maniparambil, Chris Vorster, Derek Molloy, Noel Murphy, Kevin McGuinness,
  and Noel~E. O'Connor.
\newblock Enhancing {CLIP} with {GPT-4:} harnessing visual descriptions as
  prompts.
\newblock In \emph{{IEEE/CVF} International Conference on Computer Vision,
  {ICCV} 2023 - Workshops, Paris, France, October 2-6, 2023}, pages 262--271.
  {IEEE}, 2023.

\bibitem[Menon and Vondrick(2023)]{DBLP:conf/iclr/MenonV23}
Sachit Menon and Carl Vondrick.
\newblock Visual classification via description from large language models.
\newblock In \emph{The Eleventh International Conference on Learning
  Representations, {ICLR} 2023, Kigali, Rwanda, May 1-5, 2023}. OpenReview.net,
  2023.

\bibitem[Nilsback and Zisserman(2008)]{DBLP:conf/icvgip/NilsbackZ08}
Maria{-}Elena Nilsback and Andrew Zisserman.
\newblock Automated flower classification over a large number of classes.
\newblock In \emph{Sixth Indian Conference on Computer Vision, Graphics {\&}
  Image Processing, {ICVGIP} 2008, Bhubaneswar, India, 16-19 December 2008},
  pages 722--729. {IEEE} Computer Society, 2008.

\bibitem[Oikarinen et~al.(2023)Oikarinen, Das, Nguyen, and
  Weng]{DBLP:conf/iclr/OikarinenDNW23}
Tuomas~P. Oikarinen, Subhro Das, Lam~M. Nguyen, and Tsui{-}Wei Weng.
\newblock Label-free concept bottleneck models.
\newblock In \emph{The Eleventh International Conference on Learning
  Representations, {ICLR} 2023, Kigali, Rwanda, May 1-5, 2023}. OpenReview.net,
  2023.

\bibitem[Parkhi et~al.(2012)Parkhi, Vedaldi, Zisserman, and
  Jawahar]{DBLP:conf/cvpr/ParkhiVZJ12}
Omkar~M. Parkhi, Andrea Vedaldi, Andrew Zisserman, and C.~V. Jawahar.
\newblock Cats and dogs.
\newblock In \emph{2012 {IEEE} Conference on Computer Vision and Pattern
  Recognition, Providence, RI, USA, June 16-21, 2012}, pages 3498--3505. {IEEE}
  Computer Society, 2012.

\bibitem[Pratt et~al.(2023)Pratt, Covert, Liu, and
  Farhadi]{DBLP:conf/iccv/PrattCLF23}
Sarah~M. Pratt, Ian Covert, Rosanne Liu, and Ali Farhadi.
\newblock What does a platypus look like? generating customized prompts for
  zero-shot image classification.
\newblock In \emph{{IEEE/CVF} International Conference on Computer Vision,
  {ICCV} 2023, Paris, France, October 1-6, 2023}, pages 15645--15655. {IEEE},
  2023.

\bibitem[Radford et~al.(2021)Radford, Kim, Hallacy, Ramesh, Goh, Agarwal,
  Sastry, Askell, Mishkin, Clark, Krueger, and
  Sutskever]{DBLP:conf/icml/RadfordKHRGASAM21}
Alec Radford, Jong~Wook Kim, Chris Hallacy, Aditya Ramesh, Gabriel Goh,
  Sandhini Agarwal, Girish Sastry, Amanda Askell, Pamela Mishkin, Jack Clark,
  Gretchen Krueger, and Ilya Sutskever.
\newblock Learning transferable visual models from natural language
  supervision.
\newblock In \emph{Proceedings of the 38th International Conference on Machine
  Learning, {ICML} 2021, 18-24 July 2021, Virtual Event}, pages 8748--8763.
  {PMLR}, 2021.

\bibitem[Russakovsky et~al.(2015)Russakovsky, Deng, Su, Krause, Satheesh, Ma,
  Huang, Karpathy, Khosla, Bernstein, Berg, and
  Fei{-}Fei]{DBLP:journals/ijcv/RussakovskyDSKS15}
Olga Russakovsky, Jia Deng, Hao Su, Jonathan Krause, Sanjeev Satheesh, Sean Ma,
  Zhiheng Huang, Andrej Karpathy, Aditya Khosla, Michael~S. Bernstein,
  Alexander~C. Berg, and Li Fei{-}Fei.
\newblock Imagenet large scale visual recognition challenge.
\newblock \emph{Int. J. Comput. Vis.}, 115\penalty0 (3):\penalty0 211--252,
  2015.

\bibitem[Shang et~al.(2024)Shang, Zhou, Zhang, Ni, Yang, and
  Wang]{shang2024incremental}
Chenming Shang, Shiji Zhou, Hengyuan Zhang, Xinzhe Ni, Yujiu Yang, and Yuwang
  Wang.
\newblock Incremental residual concept bottleneck models.
\newblock In \emph{Proceedings of the IEEE/CVF Conference on Computer Vision
  and Pattern Recognition}, pages 11030--11040, 2024.

\bibitem[Srivastava et~al.(2024)Srivastava, Yan, and Weng]{srivastava2024vlg}
Divyansh Srivastava, Ge Yan, and Lily Weng.
\newblock Vlg-cbm: Training concept bottleneck models with vision-language
  guidance.
\newblock \emph{Advances in Neural Information Processing Systems},
  37:\penalty0 79057--79094, 2024.

\bibitem[Touvron et~al.(2023)Touvron, Martin, Stone, Albert, Almahairi, Babaei,
  Bashlykov, Batra, Bhargava, Bhosale, Bikel, Blecher, Canton{-}Ferrer, Chen,
  Cucurull, Esiobu, Fernandes, Fu, Fu, Fuller, Gao, Goswami, Goyal, Hartshorn,
  Hosseini, Hou, Inan, Kardas, Kerkez, Khabsa, Kloumann, Korenev, Koura,
  Lachaux, Lavril, Lee, Liskovich, Lu, Mao, Martinet, Mihaylov, Mishra,
  Molybog, Nie, Poulton, Reizenstein, Rungta, Saladi, Schelten, Silva, Smith,
  Subramanian, Tan, Tang, Taylor, Williams, Kuan, Xu, Yan, Zarov, Zhang, Fan,
  Kambadur, Narang, Rodriguez, Stojnic, Edunov, and
  Scialom]{DBLP:journals/corr/abs-2307-09288}
Hugo Touvron, Louis Martin, Kevin Stone, Peter Albert, Amjad Almahairi, Yasmine
  Babaei, Nikolay Bashlykov, Soumya Batra, Prajjwal Bhargava, Shruti Bhosale,
  Dan Bikel, Lukas Blecher, Cristian Canton{-}Ferrer, Moya Chen, Guillem
  Cucurull, David Esiobu, Jude Fernandes, Jeremy Fu, Wenyin Fu, Brian Fuller,
  Cynthia Gao, Vedanuj Goswami, Naman Goyal, Anthony Hartshorn, Saghar
  Hosseini, Rui Hou, Hakan Inan, Marcin Kardas, Viktor Kerkez, Madian Khabsa,
  Isabel Kloumann, Artem Korenev, Punit~Singh Koura, Marie{-}Anne Lachaux,
  Thibaut Lavril, Jenya Lee, Diana Liskovich, Yinghai Lu, Yuning Mao, Xavier
  Martinet, Todor Mihaylov, Pushkar Mishra, Igor Molybog, Yixin Nie, Andrew
  Poulton, Jeremy Reizenstein, Rashi Rungta, Kalyan Saladi, Alan Schelten, Ruan
  Silva, Eric~Michael Smith, Ranjan Subramanian, Xiaoqing~Ellen Tan, Binh Tang,
  Ross Taylor, Adina Williams, Jian~Xiang Kuan, Puxin Xu, Zheng Yan, Iliyan
  Zarov, Yuchen Zhang, Angela Fan, Melanie Kambadur, Sharan Narang,
  Aur{\'{e}}lien Rodriguez, Robert Stojnic, Sergey Edunov, and Thomas Scialom.
\newblock Llama 2: Open foundation and fine-tuned chat models.
\newblock \emph{CoRR}, abs/2307.09288, 2023.

\bibitem[Wah et~al.(2011)Wah, Branson, Welinder, Perona, and
  Belongie]{wah2011caltech}
Catherine Wah, Steve Branson, Peter Welinder, Pietro Perona, and Serge
  Belongie.
\newblock The caltech-ucsd birds-200-2011 dataset.
\newblock 2011.

\bibitem[Wei et~al.(2022)Wei, Wang, Schuurmans, Bosma, Ichter, Xia, Chi, Le,
  and Zhou]{DBLP:conf/nips/Wei0SBIXCLZ22}
Jason Wei, Xuezhi Wang, Dale Schuurmans, Maarten Bosma, Brian Ichter, Fei Xia,
  Ed~H. Chi, Quoc~V. Le, and Denny Zhou.
\newblock Chain-of-thought prompting elicits reasoning in large language
  models.
\newblock In \emph{Advances in Neural Information Processing Systems 35: Annual
  Conference on Neural Information Processing Systems 2022, NeurIPS 2022, New
  Orleans, LA, USA, November 28 - December 9, 2022}, 2022.

\bibitem[Yan et~al.(2023)Yan, Wang, Zhong, Dong, He, Lu, Wang, Shang, and
  McAuley]{DBLP:conf/iccv/0003WZDHLWSM23}
An Yan, Yu Wang, Yiwu Zhong, Chengyu Dong, Zexue He, Yujie Lu, William~Yang
  Wang, Jingbo Shang, and Julian~J. McAuley.
\newblock Learning concise and descriptive attributes for visual recognition.
\newblock In \emph{{IEEE/CVF} International Conference on Computer Vision,
  {ICCV} 2023, Paris, France, October 1-6, 2023}, pages 3067--3077. {IEEE},
  2023.

\bibitem[Yang et~al.(2023)Yang, Panagopoulou, Zhou, Jin, Callison{-}Burch, and
  Yatskar]{DBLP:conf/cvpr/YangPZJCY23}
Yue Yang, Artemis Panagopoulou, Shenghao Zhou, Daniel Jin, Chris
  Callison{-}Burch, and Mark Yatskar.
\newblock Language in a bottle: Language model guided concept bottlenecks for
  interpretable image classification.
\newblock In \emph{{IEEE/CVF} Conference on Computer Vision and Pattern
  Recognition, {CVPR} 2023, Vancouver, BC, Canada, June 17-24, 2023}, pages
  19187--19197. {IEEE}, 2023.

\bibitem[Y{\"{u}}ksekg{\"{o}}n{\"{u}}l
  et~al.(2023)Y{\"{u}}ksekg{\"{o}}n{\"{u}}l, Wang, and
  Zou]{DBLP:conf/iclr/YuksekgonulW023}
Mert Y{\"{u}}ksekg{\"{o}}n{\"{u}}l, Maggie Wang, and James Zou.
\newblock Post-hoc concept bottleneck models.
\newblock In \emph{The Eleventh International Conference on Learning
  Representations, {ICLR} 2023, Kigali, Rwanda, May 1-5, 2023}. OpenReview.net,
  2023.

\bibitem[Yun et~al.(2023)Yun, Bhalla, Pavlick, and
  Sun]{DBLP:journals/tmlr/YunBPS23}
Tian Yun, Usha Bhalla, Ellie Pavlick, and Chen Sun.
\newblock Do vision-language pretrained models learn composable primitive
  concepts?
\newblock \emph{Trans. Mach. Learn. Res.}, 2023, 2023.

\end{thebibliography}
}

\clearpage
\setcounter{page}{1}
\maketitlesupplementary
\setcounter{section}{0}
\renewcommand\thesection{\Alph{section}}
\setcounter{table}{0}
\setcounter{figure}{0}
\setcounter{equation}{0}
\renewcommand{\thetable}{A\arabic{table}}
\renewcommand{\thefigure}{A\arabic{figure}}
\renewcommand{\theequation}{A\arabic{equation}}

\section{The implementation detail of DSS}
\subsection{The detail of summary step}
\label{sec:DSS}
\begin{figure*}[t]
	\centering
	{\includegraphics[width=\linewidth]{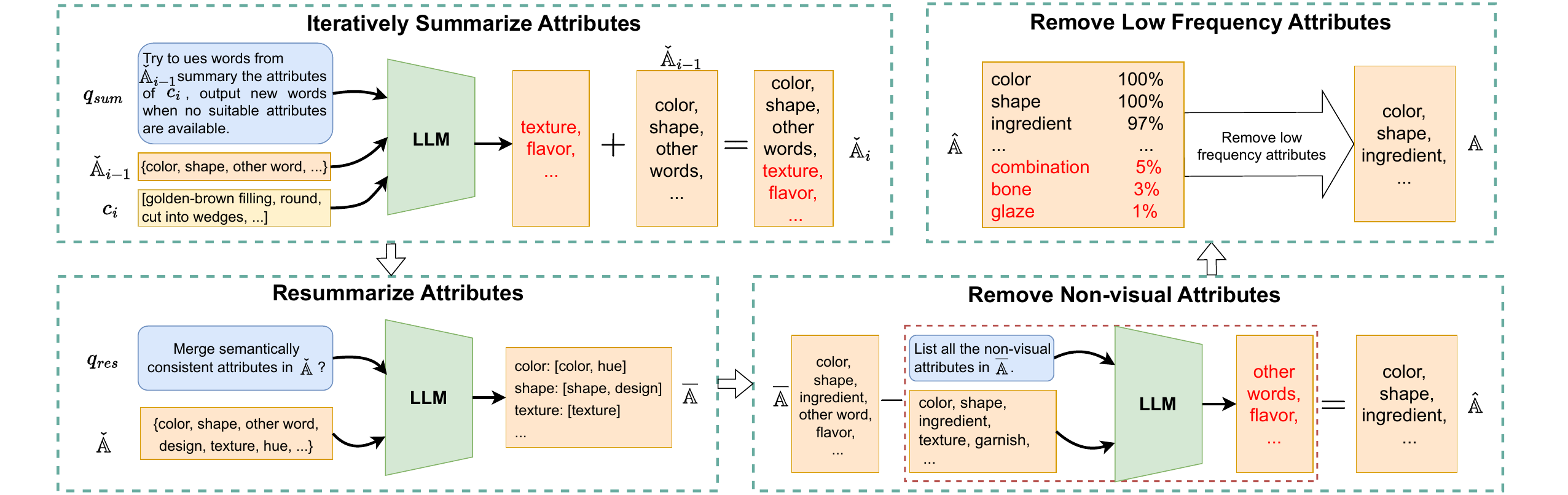}}
	\caption{Illustration of the summary step in DSS strategy. Specifically, we summarize the attributes of class concepts through the following steps: first, iteratively summarize the attributes by category; next, remove duplicate attributes from the attribute set; then, eliminate non-visual attributes; and finally, remove sparse attributes.}
	\label{fig:summary}
\end{figure*}

\begin{table*}[t]
	\resizebox{\linewidth}{!}{\begin{tabular}{ccccccccccc}
			\toprule
			& Approach         & Aircraft      & CUB           & DTD           & Flowers102    & Food101       & OxfordPets    & CIFAR-10       & CIFAR-100   &   ImageNet\\ \midrule
			Unexplainable                  & ZS-CLIP~\cite{DBLP:conf/icml/RadfordKHRGASAM21}  & 32.6          & 63.4          & 53.2          & 79.3          & 91.0          & 93.6          & \textbf{86.0} & 55.6       & 71.4  \\ \midrule
			\multirow{4}{*}{\begin{tabular}[c]{@{}c@{}}Training-free\\ Language\\ Bottleneck\end{tabular}} & VDCLIP~\cite{DBLP:conf/iclr/MenonV23}
			& -             & 63.5          & 54.4          & -             & 92.4          & 92.3          & -             & -        & 71.5     \\
			& CuPL~\cite{DBLP:conf/iccv/PrattCLF23}     & \textbf{36.7} & -             & 58.9          & 78.8          & 91.2          & 93.4          & 83.4          & 60.4      &\textbf{74.1}    \\
			& CLIP-GPT~\cite{DBLP:conf/iccvw/ManiparambilVMM23} & 34.5          & 64.8          & 56.4          & 77.8          & 91.1          & 92.8          & -             & -       &71.8      \\
			& ALBM*~(ours)     & 34.4          & \textbf{66.5} & \textbf{59.9} & \textbf{79.9} & \textbf{91.6} & \textbf{93.9} & 85.4          & \textbf{61.5} & 73.4\\ \bottomrule
	\end{tabular}}
	\caption{Comparison with zero-shot CLIP and training-free language bottlenecks on the zero-shot setting, where class names are added in the concepts, ALBM* indicate zero-shot prediction based on our collected concept sets, and ``-'' indicates that the original approaches didn't collect the concept set for the dataset.}
	\label{ZSL_with_CN}
\end{table*}

\begin{table*}[t]
	\resizebox{\linewidth}{!}{\setlength{\tabcolsep}{1.5mm}\begin{tabular}{ccccccccccc}
			\toprule
			& Approach    & Aircraft      & CUB           & DTD           & Flowers102    & Food101       & CIFAR-10      & CIFAR-100   & ImageNet  & Average       \\ \midrule
			\multirow{2}{*}{\begin{tabular}[c]{@{}c@{}}Class-Shared\\ Concept Space\end{tabular}}   & Labo~\cite{DBLP:conf/cvpr/YangPZJCY23}        & 45.6          & 78.2          & 67.6          & 92.6          & 87.6          & 85.7          & 45.5          & 71.0 & 71.7          \\
			& ALBM (ours) & \textbf{53.5} & \textbf{83.4} & \textbf{68.6} & \textbf{98.1} & \textbf{89.1} & \textbf{86.0} & \textbf{62.4} & \textbf{76.5} & \textbf{77.2}  \\ \midrule
			\multirow{2}{*}{\begin{tabular}[c]{@{}c@{}}Class-Specific\\ Concept Space\end{tabular}} & Labo~\cite{DBLP:conf/cvpr/YangPZJCY23}        & \textbf{41.2} & 69.5          & 66.3          & \textbf{95.7} & 82.9          & 80.9          & 49.5      & 69.2    & 69.4          \\
			& ALBM (ours) & 40.0          & \textbf{69.7} & \textbf{69.4} & 92.4          & \textbf{84.8} & \textbf{84.2} & \textbf{52.5} & \textbf{70.0} & \textbf{70.4} \\ \bottomrule
	\end{tabular}}
	\caption{Comparison with existing LBM approach LaBo~\cite{DBLP:conf/cvpr/YangPZJCY23} in class-shared concept space and class-specific concept space under 16-shot few-shot learning setting. For fair comparison, we use CLIP's original visual representation instead of the feature of visual attribute prompt for our approach.}
	\vspace{-3mm}
	\label{few}
\end{table*}

This subsection introduces the implementation detail of the summary step in DSS. The diagram of the summary step is shown in Fig.~\ref{fig:summary}. Considering the token length limitation of LLM and its unreliability in generating long texts, it is unable to directly ask LLM to summary the entire attribute set by giving LLM all the generated concepts $\left\{{\bm{c}}_i\right\}_{i=1}^K$ in the description step. Therefore, we query the attributes on a per-class basis. Furthermore, since quarrying LLM multiple times may lead to inconsistent outputs (e.g., nose \& snout), we adopt an iterative approach to summarize the attribute set $\check{\sA}$. That is, for each query, we encourage the LLM to use words from the existing attribute set $\check{\sA}_{i-1}$ to summarize the attributes of ${\bm{c}}_i$, and only output new words when no suitable attributes are available in $\check{\sA}_{i-1}$ based on the prompt $q_{sum}$, formally,
\begin{equation}
	\check{\sA}_i = \check{\sA}_{i-1} + {\rm LLM}\left( {\bm{c}}_i, \check{\sA}_{i-1}, q_{sum} \right).
\end{equation}
The overall attribute set $\check{\sA}$ is equal to $\check{\sA}_K$.

To further avoid duplicate and synonymous attributes, we use LLM to resummarize the attribute set with the prompt $q_{res}$, formally,

\begin{equation}
	\overline{\sA} = {\rm LLM}\left( \check{\sA}, q_{res} \right).
\end{equation}

However, the attribute set summarized by the above prompt still has two limitations. The first is that some collected attributes describe the non-visual information of classes (e.g., the alternative name, smell, etc.). Predictions based on these non-visual attributes will disturb the interpretability of the inference process. And the other limitation is that some attributes are very sparse on categories, with only a few classes having corresponding descriptions. To address the above two limitations, we first use LLM to filter non-visual attributes using the prompt $p_{vis}$, formally, 
\begin{equation}
	\hat{\sA} = \overline{\sA} - {\rm LLM}\left( \overline{\sA}, q_{vis} \right),
\end{equation}
where $\hat{\sA}$ represents the visual attribute set, ${\rm LLM}\left( \overline{\sA}, q_{vis} \right)$ is the non-visual attribute set summarized by LLM.

Then we count the number of descriptions corresponding to each attribute and remove attributes that occur with a frequency of less than $r\%$ across all classes to obtain the final attribute set $\sA$.

\subsection{The prompts used in DSS}
 This subsection introduces the prompts used in DSS. Since we directly use the concept sets collected by existing work~\cite{DBLP:conf/cvpr/YangPZJCY23} as the output of the Description Step, the prompts used in this step are identical to its released prompts. In the Summary Step, we first use $q_{sum}$ to prompt LLM iteratively summarize the attributes by category, which is as follows:
\begin{quote}
Your task is to extract attributes of different categories from the descriptions I gave you.
\\
Specially, you can complete the task by following the instructions:
\\
1. You can select the noun related to the attribute form {exsit attribute set}, and if you think the attribute describe by the phrase is not among them, you can answer other words.
\\
2. Each phrase corresponds to a description, and the number of the two should also be consistent.
\\
3. Output a Python dictionary with the \{attribute name\} as the key, and no newline required between each description. PLEASE USE ``:'' AFTER the KEY.
\end{quote}

Subsequently, we use LLM to remove duplicate attributes from the attribute set with the prompt $q_{res}$:
\begin{quote}
	Your task is to merge the attributes I give you into semantically consistent attribute groups.
	\\
	Specially, you can complete the task by following the instructions:
	\\
	1. Only merge the attributes I give, and only merge semantically consistent attributes.
	\\
	2. The semantics of the merged attributes should not be repeated.
	\\
	3. The words representing an attribute group must be the words of the attributes I give, and the words in the same attribute group must all come from the attributes I give.
	\\
	4. The sum of the words in all attribute groups should be equal to the attribute set I gave.
	\\
	5. Output some python lists, each list represents a attribute group.
	\\
	===
	\\
	Please merge semantically consistent attribute among the attributes {attribute set}:
	\\
	===
\end{quote}

Next, we use LLM to remove non-visual attributes with the prompt $q_{vis}$:
\begin{quote}
	Suppose you have some photos of \{all class name\}, please write down \{attribute set\} in order whether these attributes are the visual attributes of these pictures:
\end{quote}

In the Supplement Step, we utilize LLM to supply the missing concepts with the prompt $q_{sup}$ as follows:
\begin{quote}
	Your task is to describe a certain attribute of a certain class.
	\\
	Specially, you can complete the task by using short and precise descriptions. And no newline is required before each description.
	\\
	===
	\\
	Please describe the attribute \{attribute\} of the class \{class name\} according to the following examples, and no newline required between each description:
	\\
	===
\end{quote}
where the content inside the curly braces represents the corresponding variables.

\section{Additional analysis}
\subsection{Zero-shot performance with class names}
In Tab.~\ref{ZSL}, we compared our approach with existing TfLB approaches under the setting where class descriptions only include visual concepts without class names, to rigorously evaluate the performance of interpretable image recognition. However, the performance of existing TfLB approaches suffers significantly under this setting. As a result, existing TfLB approaches~\cite{DBLP:conf/iccv/0003WZDHLWSM23, DBLP:conf/iccv/PrattCLF23, DBLP:conf/iccvw/ManiparambilVMM23} recommend including class names in the descriptions, such as "a photo of a {class name}, which has/is {class concept}," to achieve better classification performance.
To further validate the effectiveness of our proposed method, we conducted comparative experiments under this setting as well, as shown in Tab.~\ref{ZSL_with_CN}. From Tab.~\ref{ZSL_with_CN}, it can be seen that our approach achieves the best performance on 6 out of 9 datasets, slightly underperforming the current state-of-the-art results on only 3 datasets. These results demonstrate the effectiveness of our proposed DSS strategy, which extracts more comprehensive visual information for each class by summarizing a cross-class shared attribute set.

\subsection{Comparison with existing LBM in class-shared and class-specific concept space}
In Section~4.3, we analyzed the reason for our relatively worse performance on the base classes in the Aircraft and Food101 datasets compared with existing LBMs is that existing LBMs learn in a category-shared concept space, where they exploit explainable spurious cues to achieve better performance. To further verify this, we compared our ALBM with the existing LBM approach LaBo~\cite{DBLP:conf/cvpr/YangPZJCY23} in both class-shared concept space (where the concept classifier identifies classes based on concepts from all classes) and class-specific concept space (where the concept classifier identifies classes based on concepts specific to them), as shown in Tab.~\ref{few}. It is worth noting that, for a fair comparison, we do not use visual attribute prompts here but instead use CLIP's original visual representation. Additionally, CLBM is not applicable to the category-specific concept space because its concept set does not provide a mapping between concepts and categories. From Tab.~\ref{few}, it is clear that, in general, ALBM outperforms existing LBM methods in both class-shared and class-specific settings, demonstrating that the concept set we generate with a unified attribute set better reflects the visual information of classes. Furthermore, the performance of both LaBo and ALBM in the category-specific concept space is weaker than in the category-shared concept space, which highlights the trade-off between interpretability and performance. This is due to the insufficient interpretability of features extracted by the CLIP model. Therefore, we further propose VAPL to extract features on each fine-grade attribute.

\begin{figure*}[t]
	\centering
	\begin{subfigure}[b]{0.24\textwidth}
		\centering
		\includegraphics[width=\textwidth]{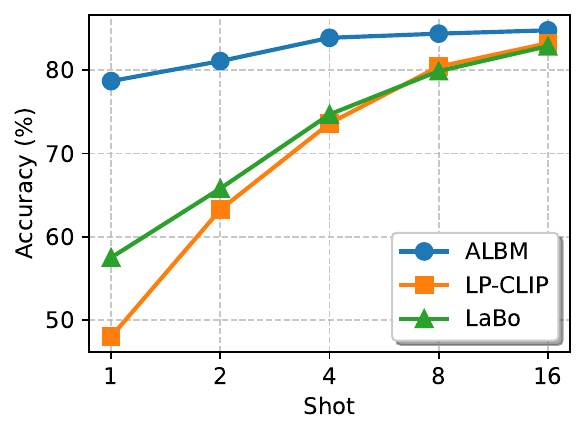}
		\caption{Food101}
		\label{fig:sub1}
	\end{subfigure}
	\hfill
	\begin{subfigure}[b]{0.24\textwidth}
		\centering
		\includegraphics[width=\textwidth]{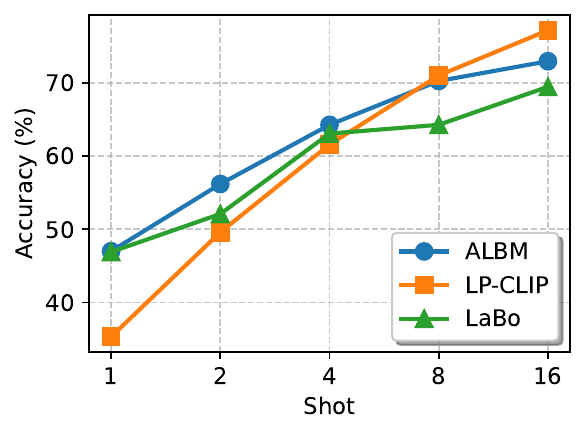}
		\caption{CUB}
		\label{fig:sub2}
	\end{subfigure}
	\hfill
	\begin{subfigure}[b]{0.24\textwidth}
		\centering
		\includegraphics[width=\textwidth]{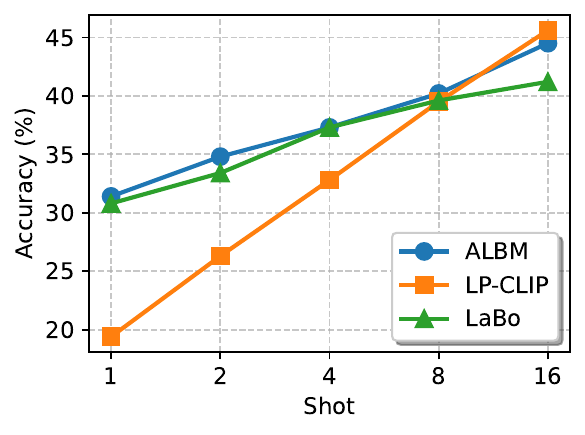}
		\caption{Aircraft}
		\label{fig:sub3}
	\end{subfigure}
	\hfill
	\begin{subfigure}[b]{0.24\textwidth}
		\centering
		\includegraphics[width=\textwidth]{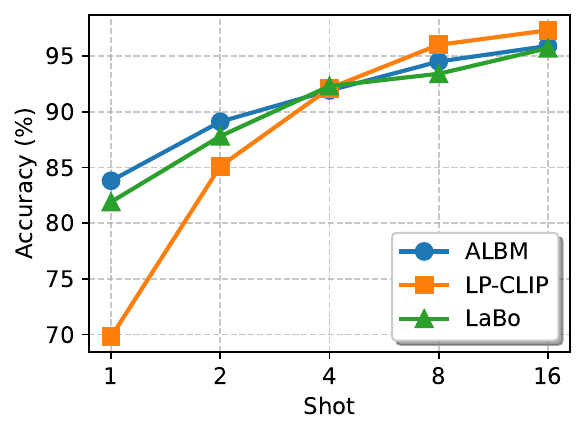}
		\caption{Flowers102}
		\label{fig:sub4}
	\end{subfigure}
	\caption{Few-shot performance comparison between our ALBM, LP-CLIP~\cite{DBLP:conf/icml/RadfordKHRGASAM21}, and LaBo~\cite{DBLP:conf/cvpr/YangPZJCY23} on Food101, CUB, Aircraft, and Flowers102 datasets.}
	\label{fig:few-shot}
\end{figure*}

\begin{figure*}[th]
	\centering
	\begin{subfigure}[b]{0.24\textwidth}
		\centering
		\includegraphics[width=\textwidth]{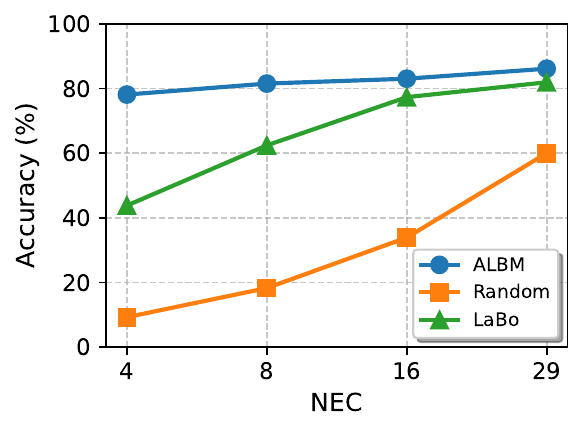}
		\caption{Food101}
		\label{fig:NEC1}
	\end{subfigure}
	\hfill
	\begin{subfigure}[b]{0.24\textwidth}
		\centering
		\includegraphics[width=\textwidth]{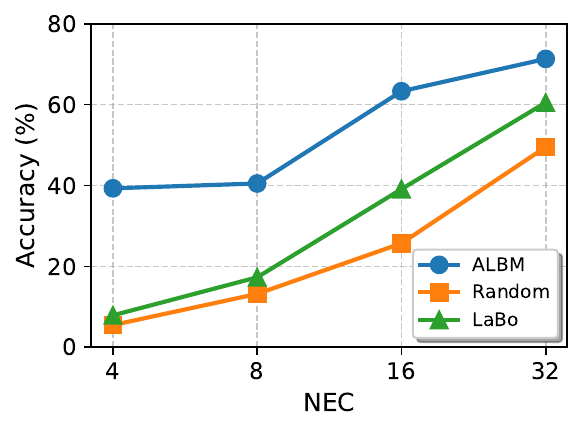}
		\caption{CUB}
		\label{fig:NEC2}
	\end{subfigure}
	\hfill
	\begin{subfigure}[b]{0.24\textwidth}
		\centering
		\includegraphics[width=\textwidth]{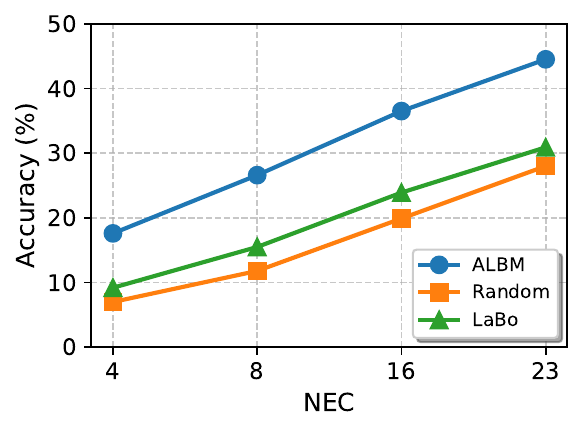}
		\caption{Aircraft}
		\label{fig:NEC3}
	\end{subfigure}
	\hfill
	\begin{subfigure}[b]{0.24\textwidth}
		\centering
		\includegraphics[width=\textwidth]{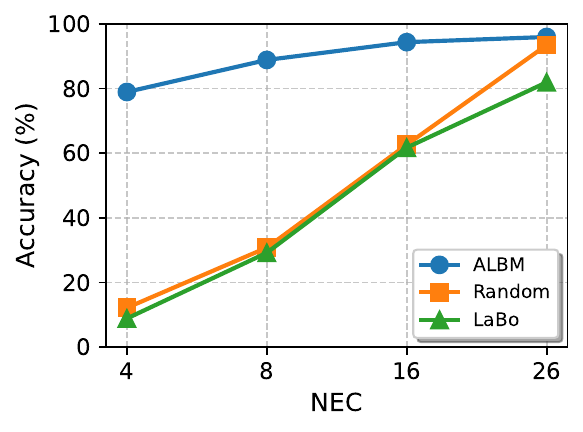}
		\caption{Flowers102}
		\label{fig:NEC4}
	\end{subfigure}
	\caption{16-shot performance comparison between our ALBM,  LaBo~\cite{DBLP:conf/cvpr/YangPZJCY23}, and randomly initialized concept bottleneck layer under different NECs. The experiments are conducted on Food101, CUB, Aircraft, and Flowers102 datasets.}
	\label{fig:NEC}
\end{figure*}

\begin{table*}[th]
	\centering
	\resizebox{\linewidth}{!}{\setlength{\tabcolsep}{1.5mm}	\begin{tabular}{ccccccccccc}
			\toprule
			CLIP Version & Aircraft      & CUB           & DTD           & Flowers102    & Food101       & OxfordPets    & CIFAR-10       & CIFAR-100  & ImageNet & Average    \\ \midrule
			ViT-B/32 (88M)     & 12.5          & 16.9          & 38.9          & 30.3          & 56.4          & 28.5          & 66.6          & 30.6    & 51.0 &   38.1    \\
			ViT-B/16 (88M)    & 14.3          & 17.2          & 40.7          & 43.0          & 58.8          & 32.0          & 79.0          & 33.4     & 55.5 &    41.5  \\
			ViT-L/14 (304M)    & \textbf{18.0} & \textbf{25.0} & \textbf{48.5} & \textbf{54.9} & \textbf{75.4} & \textbf{35.9} & \textbf{83.1} & \textbf{43.1} & \textbf{64.6} & \textbf{49.8} \\ \bottomrule
	\end{tabular}}
	\caption{Zero-shot classification performance of ALBM with different CLIP versions, where the values in parentheses represent the parameter size of model.}
	\label{ZSL_clip}
	\vspace{-3mm}
\end{table*}

\begin{table*}[th]
	\resizebox{\linewidth}{!}{\setlength{\tabcolsep}{0.7mm}\begin{tabular}{ccccccccccccccccccccc}
			\toprule
			\multirow{2}{*}{\begin{tabular}[c]{@{}c@{}}CLIP\\ Version\end{tabular}} & \multicolumn{2}{c}{Aircraft}  & \multicolumn{2}{c}{CUB}       & \multicolumn{2}{c}{DTD}       & \multicolumn{2}{c}{Flowers102} & \multicolumn{2}{c}{Food101}   & \multicolumn{2}{c}{OxfordPets} & \multicolumn{2}{c}{CIFAR-10}   & \multicolumn{2}{c}{CIFAR-100} & \multicolumn{2}{c}{ImageNet} & \multicolumn{2}{c}{Average} \\ \cmidrule{2-21}
			& Base          & Novel         & Base          & Novel         & Base          & Novel         & Base           & Novel         & Base          & Novel         & Base           & Novel         & Base          & Novel         & Base          & Novel   & Base          & Novel  & Base          & Novel    \\ \midrule
			ViT-B/32                                                                & 22.7          & 22.2          & 60.8          & 20.0          & 71.9          & 52.7          & 84.4           & 25.0          & 72.1          & 72.4          & 63.8           & 45.2          & 79.7          & 73.5          & 44.9          & 37.9  & 61.0 & 59.8   & 62.3 &  45.4    \\
			ViT-B/16                                                                & 30.3          & 25.4          & 61.5          & 22.0          & 75.0          & 55.7          & 88.1           & 26.5          & 78.6          & 78.6          & 69.1           & 54.0          & 81.0          & 86.9          & 48.6          & 37.5  & 68.2 & 67.3   & 66.7 &  50.4   \\
			ViT-L/14                                                                & \textbf{38.7} & \textbf{33.0} & \textbf{91.9} & \textbf{27.8} & \textbf{78.6} & \textbf{60.5} & \textbf{91.7}  & \textbf{32.4} & \textbf{88.5} & \textbf{86.8} & \textbf{79.2}  & \textbf{61.1} & \textbf{90.8} & \textbf{93.6} & \textbf{59.3} & \textbf{55.1} & \textbf{75.0} & \textbf{73.9}& \textbf{77.0} & \textbf{58.3} \\ \bottomrule
	\end{tabular}}
	\caption{Base-to-novel classification performance of ALBM with different CLIP versions.}
	\vspace{-3mm}
	\label{B2N_clip}
\end{table*}

\subsection{Few-shot performance comparison}
Yang et al.~\cite{DBLP:conf/cvpr/YangPZJCY23} found that compared to linear-probe CLIP, LBM achieves better few-shot performance by incorporating class concept information, which enhances the image recognition process. To evaluate the few-shot capability of our approach, we compare its performance with LaBo and LP-CLIP on Food101, CUB, Aircraft, and Flowers102 datasets, as shown in Fig.~\ref{fig:few-shot}. It is clear that compared to LP-CLIP, ALBM demonstrates significant performance advantages, particularly when the number of training samples is extremely low. Additionally, it outperforms LaBo, further emphasizing its effectiveness. These results highlight that our collection strategy enhances few-shot learning by introducing more informative class concepts.

\subsection{Interpretability verification via sparse prediction}
To further verify the interpretability of ALBM, we evaluated the accuracy under diifferent NECs. Number of Effective Concepts (NEC)~\cite{srivastava2024vlg} is a newly proposed CBM interpretability metric that restricts the number of concepts with nonzero weights, which is motivated by the observation that when the number of concepts is very large, even those lacking interpretability can achieve high performance. Conversely, when concepts are sparse, only interpretable ones can provide sufficient information for recognition. Thus, by limiting NEC, different LBMs can be fairly compared in terms of interpretability and performance. As shown in Fig.~\ref{fig:NEC}, our approach achieves superior performances compared with LaBo and random concepts, further verifying the interpretability of ALBM.

\subsection{Relationship between interpretability and model size}
In this subsection, we further analyze the relationship between interpretability and model size to provide guidance on model selection for the user of interpretable classification models. Therefore, we compare the zero-shot and base-to-novel classification performance of ALBM models using different versions of CLIP, as shown in Tab.~\ref{ZSL_clip} \& \ref{B2N_clip}. By comparing the first and second rows of Tab.~\ref{ZSL_clip} \& \ref{B2N_clip}, it can be observed that ViT-B/16 outperforms ViT-B/32 in all settings. This is due to that although these CLIP models have the same number of parameters, the smaller patches in the ViT-B/16 version preserve more local features, resulting in stronger interpretability.
Furthermore, as shown in the third row of Tab.~\ref{ZSL_clip} \& \ref{B2N_clip}, CLIP with larger parameter size consistently outperforms its smaller versions, demonstrating a significant improvement in the ability to capture interpretable fine-grained attribute features. This aligns with the scaling law. Therefore, we recommend using larger models for interpretable image recognition whenever resources allow.

\end{document}